\documentclass[nohyperref]{article}
\usepackage{microtype}
\usepackage{graphicx}
\usepackage{booktabs} % for professional tables
\usepackage{url}
\usepackage{graphicx}	% For figure environment
\usepackage{hyperref}
\usepackage{appendix}

\usepackage[utf8]{inputenc} % allow utf-8 input
\usepackage[T1]{fontenc}    % use 8-bit T1 fonts
\usepackage{hyperref}       % hyperlinks
\usepackage{url}            % simple URL typesetting
\usepackage{booktabs}       % professional-quality tables
\usepackage{amsfonts}       % blackboard math symbols
\usepackage{nicefrac}       % compact symbols for 1/2, etc.
\usepackage{microtype}      % microtypography
\usepackage{xcolor}         % colors
\usepackage{graphicx}
\usepackage{wrapfig}
\usepackage{gensymb}
\usepackage{wrapfig}
\usepackage{pdfpages}
\usepackage{float}
\usepackage{subcaption}
\usepackage{tabularx}
\usepackage{bbm}

\usepackage{paralist}

\usepackage{parskip}
\usepackage{kantlipsum}

\newcommand{\specialcell}[2][c]{%
\begin{tabular}[#1]{@{}c@{}}#2\end{tabular}}

\newcommand*{\crosssymbol}{%
    \text{%
      \raise 1ex\hbox{%
        \rlap{\vrule height.2pt depth.2pt width .75ex}%
        \hbox to .75ex{\hss\vrule height .5ex depth 1ex\hss}%
      }%
    }%  
}

\usepackage{hyperref}

\usepackage[accepted]{icml2022}

\usepackage{amsmath}
\usepackage{amssymb}
\usepackage{mathtools}
\usepackage{amsthm}

\usepackage[capitalize,noabbrev]{cleveref}

\theoremstyle{plain}

\theoremstyle{definition}

\theoremstyle{remark}

\usepackage[textsize=tiny]{todonotes}

\icmltitlerunning{DETRtime}

\begin{document}

\twocolumn[
\icmltitle{\textit{A Deep Learning Approach for the Segmentation of\\ Electroencephalography Data in Eye Tracking Applications}}



\icmlsetsymbol{equal}{*}
\icmlsetsymbol{tech}{\crosssymbol}

\begin{icmlauthorlist}
\icmlauthor{Lukas Wolf}{eth0,uzh}
\icmlauthor{Ard Kastrati}{eth}
\icmlauthor{Martyna Beata P{\l}omecka}{uzh}
\icmlauthor{Jie-Ming Li}{eth0}
\icmlauthor{Dustin Klebe}{eth0}
\icmlauthor{Alexander Veicht}{eth0}
\icmlauthor{Roger Wattenhofer}{eth}
\icmlauthor{Nicolas Langer}{uzh}
\end{icmlauthorlist}

% \textsuperscript{*}Equal contribution
\icmlaffiliation{eth0}{Department of Computer Science, ETH Zurich, Zurich, Switzerland}
\icmlaffiliation{eth}{Information Technology Department, ETH Zurich, Zurich, Switzerland}
\icmlaffiliation{uzh}{Department of Psychology, University of Zurich, Zurich, Switzerland}

%\icmlaffiliation{comp}{Company Name, Location, Country}
%\icmlEqualContribution{as}

\icmlcorrespondingauthor{Lukas Wolf}{wolflu@ethz.ch}
\icmlcorrespondingauthor{Ard Kastrati}{kard@ethz.ch}
\icmlcorrespondingauthor{Martyna Plomecka}{martyna.plomecka@uzh.ch}

\icmlkeywords{Machine Learning, ICML}

\vskip 0.3in]


\printAffiliationsAndNotice{}  % leave blank if no need to mention equal contribution

\begin{abstract}
%The collection of eye gaze information provides a window into many critical aspects of human cognition, health and behaviour. Additionally, many neuroscientific studies complement the behavioural information gained from eye tracking with the high temporal resolution and neurophysiological markers provided by electroencephalography (EEG). One of the essential eye-tracking software processing steps is the segmentation of the continuous data stream into events relevant to eye-tracking applications, such as saccades, fixations, and blinks. Here, we introduce DETRtime, a novel framework for time-series segmentation that creates ocular event detectors that do not require additionally recorded eye-tracking modality and rely solely on EEG data. Our end-to-end deep-learning-based framework brings recent advances in Computer Vision to the forefront of the times series segmentation of EEG data. The framework was trained and tested on simultaneously recorded, high-density EEG and eye-tracking data from 80 participants. We successfully decomposed the continuous EEG data stream into ocular events outperforming current state-of-the-art deep learning models, achieving F1 score above 0.9. This work is a step toward opening new avenues of research and opportunities for a broader segment of the EEG and Eye-Tracking community.
The collection of eye gaze information provides a window into many critical aspects of human cognition, health and behaviour. Additionally, many neuroscientific studies complement the behavioural information gained from eye tracking with the high temporal resolution and neurophysiological markers provided by electroencephalography (EEG). One of the essential eye-tracking software processing steps is the segmentation of the continuous data stream into events relevant to eye-tracking applications, such as saccades, fixations, and blinks. Here, we introduce DETRtime, a novel framework for time-series segmentation that creates ocular event detectors that do not require additionally recorded eye-tracking modality and rely solely on EEG data. Our end-to-end deep-learning-based framework brings recent advances in Computer Vision to the forefront of the times series segmentation of EEG data. \mbox{DETRtime} achieves state-of-the-art performance in ocular event detection across diverse eye-tracking experiment paradigms. In addition to that, we provide evidence that our model generalizes well in the task of EEG sleep stage segmentation. 
\end{abstract}

\section{Introduction and Motivation}
\label{sec:introduction-motivation}

Eye gaze information is widely used in cognitive neuroscience and psychology. Moreover, many neuroscientific studies complement scientific methods such as functional magnetic resonance imaging (fMRI) and electroencephalography (EEG) with eye-tracking technology to identify variations in attention, arousal, and the participant's compliance with the task demands \cite{hanke2019practical}. This way, the combined EEG and eye-tracking studies can provide an ``ideal neuroscience model'' to investigate brain-behaviour associations \cite{luna2008development}. Nowadays, infrared video-based eye trackers are the most common approach in scientific studies. Although costs for infrared video-based are slowly becoming less prohibitive, these devices remain outside the range of access for many researchers due to the added layers of complexity (e.g., operator training, setup time, synchronization of eye-tracking data with fMRI or EEG, and analysis of an additional data type) that can be burdensome \cite{holmqvist2011eye}. In addition, once installed, setup and calibration for each scanning session also can be time-consuming \cite{fuhl2016pupil}.
\\
%For that reason, the  ever growing body of neuroscientists started moving towards the 
%the Machine learning techniques that allow extracting information from EEG recordings of brain activity and play a crucial role in several significant EEG‐based research and application areas \cite{roy}. In particular, deep learning allows computational models to learn representations of data with multiple abstraction levels and, therefore, to use all of the information that the dataset has to offer \cite{nat}. 
Nonetheless, recent evidence \cite{10.1145/3448018.3458014, DBLP:journals/corr/abs-2111-05100} suggests that gaze position can be computed by using the combination of EEG and deep learning. Our benchmark and a dataset for estimation of gaze position intending to simulate a purely EEG-based eye-tracker shows that this task is challenging to solve with high accuracy.\\
However, even the segmentation of continuous EEG data stream into ocular events (i.e. fixations, saccades and blinks), the prerequisite for an EEG based eye-tracker, can provide meaningful insights into human behaviour \cite{toivanen2015probabilistic}. For instance, blinks can indicate various human states, including fatigue or drowsiness \cite{schmidt2018eye}. Moreover, the analysis of the gaze activity recognition allows us to study changes in cognitive workload
\cite{buettner2013cognitive}, recognize the context from gaze pattern behaviour \cite{bulling2011recognition}, and even detect microsleep episodes in eye movement data while driving a car \cite{behrens2010improved}. Despite these benefits, no deep-learning based approach that segments the EEG signal into meaningful ocular events is available in daily research and clinical practice. 
%production-ready eye-tracking system contains at least three components: the segmentation of the continuous data stream (into fixations, saccades and blinks), the sample creation process from this stream, and finally, the actual estimation of the participant's gaze position (on the screen).
%Although \cite{DBLP:journals/corr/abs-2111-05100} provides a baseline for the 
%estimation of gaze position, 
\\
To address this shortcoming, we developed the first neural network-based segmentation method to identify and differentiate ocular events from continuous EEG data. In this paper, we introduce DETRtime, a novel time series network architecture that can be used for the segmentation of events in a unified manner. The DETRtime is based on the DETR architecture \cite{detr} that was initially proposed for object detection. 
\\
Our time series adaptation maps sequential inputs of arbitrary length to sequences of class labels on a freely chosen temporal scale. 
We evaluated DETRtime on EEG data consisting of recordings from 4 experimental paradigms and overall 168 participants. In addition, we provided evidence on a second task, namely EEG sleep stage segmentation. For that matter, we ran experiments on the publicly available \textit{Sleep-EDF-153} dataset \cite{sleep-edf}.

In all cases, we found that DETRtime outperforms current state-of-the-art deep learning models on both the ocular event and sleep stage segmentation. 
\\
To conclude, our main contributions are:

\begin{itemize}
    \item A novel, purely EEG-based segmentation technique for eye-tracking applications.
%    \item A Conditional Random Field (CRF)-based U-Net segmentation architecture related to U-Time \cite{perslev2019utime}.
    \item An adaption of the Detection Transformer (DETR) architecture \cite{detr}, introducing \textit{instance segmentation} to the analysis of EEG data and providing a performant alternative to the current state-of-the-art which implements (often convolution-based) \textit{semantic segmentation} \cite{DBLP:journals/corr/Thoma16a}.
    %\item A unique large dataset, containing 108.4K fixations, 108.5K saccades and 10.2K blinks,  with concurrently recorded EEG and infrared video-based eye tracking  serving as ground truth.
    \item A collection of two novel ocular event datasets (Movie Watching and Large Grid paradigm) containing 139k fixations, 139k saccades and 14k blinks, with concurrently recorded EEG and infrared video-based eye-tracking serving as ground truth.  
    %In addition to that, we provide evidence on two additional, publicly available ocular event datasets as well as on the task of EEG sleep stage segmentation .
\end{itemize}

\section{Related Work}  
\label{sec:related-work}
%In the past decade, many research labs have combined EEG and eye-tracking recordings
%Research in computer vision has traditionally been at the forefront of this work .  
%Moreover, ocular events' detection is an active research topic with applications in human behaviour analysis \cite{mele2012gaze}, activity recognition \cite{bulling2011recognition}, human-computer interaction and usability research \cite{jacob2003eye}, to mention a few.
The ocular events' detection is an active research topic with applications in human behaviour analysis \cite{mele2012gaze}, activity recognition \cite{bulling2011recognition}, human-computer interaction and usability research \cite{jacob2003eye}, to mention a few.
\\
The most standard and exploited segmentation techniques rely on infrared video-based systems. Modern eye-tracker companies use proprietary solutions to enable fast and reliable data segmentation \cite{engbert2003microsaccades, nystrom2010adaptive}. These approaches utilize adaptive thresholds freeing researchers from setting different thresholds per trial when the noise level varies between trials. Another notable example of the segmentation framework was provided by \cite{zemblys2018gazeNet}. This deep learning-based solution detects eye movements from eye-tracking data and does not require hand-crafted signal features or signal thresholding. However, in many experimental settings, a camera setup for eye tracking is not available, and thus this approach becomes impossible.
\\
Another technique of measuring eye movements is Electrooculography (EOG), which records changes in electric potentials that originate from movements of the eye muscles \cite{barea2002system}. Previous studies have described and evaluated algorithms for detecting saccade, fixation, and blink characteristics from EOG signals recorded while driving a car \cite{behrens2010improved}. The proposed algorithm detects microsleep episodes in eye movement data, showing the high importance of the tool.
Other authors \cite{bulling2009eye} have successfully
evaluated algorithms for detecting three eye movement types from EOG signal achieving average precision of 76.1 \% and recall of 70.5 \% over all classes and participants. 
To date, the most successful and comprehensive approach to the problem of the EOG segmentation was proposed by \cite{pettersson2013algorithm}. They classified temporal eye parameters (saccades, blinks) from EOG data. The classification sensitivity for horizontal and large saccades was larger than 89\% and for vertical saccades larger than 82\%.
Another line of research on the subject has been mostly restricted to blink detection \cite{kleifges2017blinker}. Their BLINKER algorithm effectively captures most of the blinks and calculates common ocular indices. 
Nonetheless, to the best of our knowledge, an end-to-end continuous EEG data stream segmentation framework for eye-tracking applications is missing.
One of the prominent attempts for EEG data segmentation applied to sleep staging was targeted in \cite{perslev2019utime}. The proposed architecture, U-Time, could learn sleep staging based on various input channels across healthy and diseased subjects, obtaining a global F1 score above 0.75. 
However, the authors did not address the problem of segmenting the continuous EEG data stream into ocular events. Nevertheless, they recommended their network architecture as a universal tool for the segmentation of various psychophysiological datasets \cite{perslev2019utime}. Recent improvements in sleep staging research were obtained by the introduction of SalientSleepNet \cite{salientsleepnet}. The proposed fully convolutional network is based on the $U^2$-Net architecture, originally proposed for salient object detection in computer vision. SalientSleepNet shows performance improvements over U-Time on the publicly available Sleep-EDF-39, and Sleep-EDF-153 datasets \cite{sleep-edf}. Thus, we apply both U-Time and SalientSleepNet as part of our baseline efforts.

~%Finally, previous work shows the feasibility of using machine and deep learning models that combine the two modalities EEG and Eye-Tracking. \cite{emotionmeter,liu2022identifying,eegetmultimodal} apply deep learning in recognition of emotions by combining EEG and Eye-tracking data.

%Existing event detection algorithms for eye-movement data almost exclusively rely on thresholding one or more hand-crafted signal features, each computed from the stream of raw gaze data. Segmentation of eye-movement data was already properly solved in \cite{zemblys2018gazeNet}. However, in many experimental settings, a camera set up for eye tracking is not available, and thus this approach becomes impossible.
%\\
%Up until now, a purely EEG-based approach with high accuracy to segment a continuous EEG data stream for eye-tracking applications is missing. A similar goal, namely sleep stage segmentation, was targeted in \cite{perslev2019utime}. As recurrent models are difficult to tune and optimize, a novel \textit{U-Time} model was proposed, based on the U-Net architecture, which was initially presented for image segmentation. U-Time maps sequential inputs of arbitrary length to sequences of class labels.
%In \cite{segmentation_comparison}, a variety of algorithms for 1D signal segmentations are discussed. Algorithms selected for this purpose share useful common features, like adaptive and sequential (not block-wise) processing, making them good candidates for real-time time-series processing.

%\iffalse

%\section{Methods}
%\label{sec:methods}

\section{Experimental Setup and Dataset}
\label{sec:setup-data}
The experiment took place in a sound-attenuated and darkened room. The participant was seated at a distance of 68cm from a 24-inch monitor (ASUS ROG, Swift PG248Q, display dimensions 531×299 mm, resolution 800×600 pixels resulting in a display: 400×298.9 mm, vertical refresh rate of 100Hz). Participants completed the tasks sitting alone, while research assistants monitored their progress in the adjoining room. The study was approved by the local ethics commission.

%The study was approved by the Ethics Commission of the University of Zurich. 

\subsection{Data Acquisition}

\label{sec:acquisition}

\begin{figure}[t]
    \centering
    \includegraphics[width=0.45\textwidth]{ 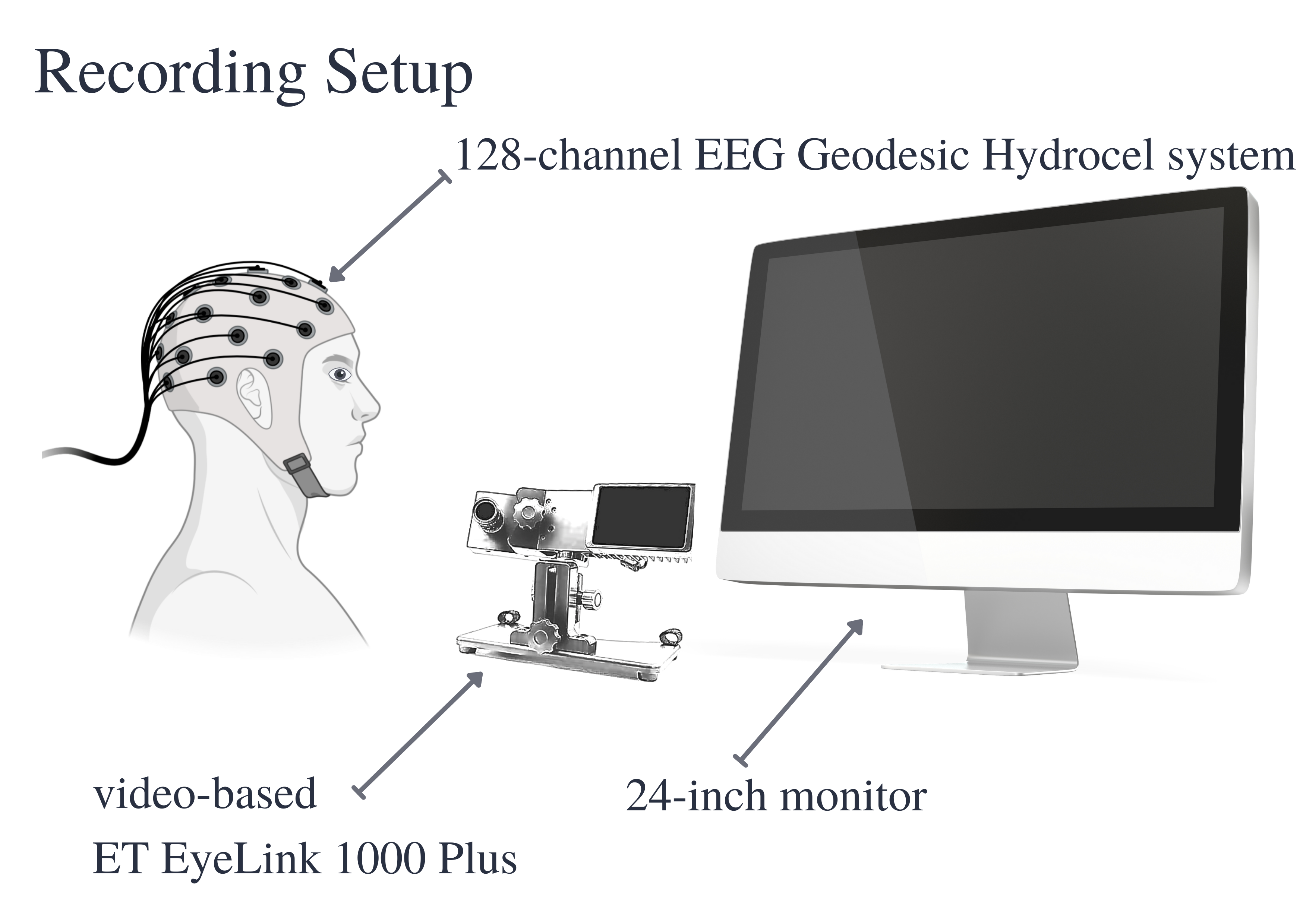}

    \caption{The illustration of the recording setup}
    %The target value (in training) and the ideal prediction (in testing) of our model for the 4-th cell would look like: (0.45, 0.136, 1, 1, 0) assuming that the red object belongs to class 1 (saccade)
    \label{fig:eeg}
    %\vspace{-0.5cm}
\end{figure}

\paragraph{Participants}
We collected data from 168 healthy adults (age range 20-80 years, 92 females) across four experimental paradigms (see Section \ref{sec:stimuli-and-experimental-design}). All participants gave written consent for their participation and the re-use of the data prior to the start of the experiments. All participants received a monetary compensation (the local currency equivalent of 25 United States Dollars) per hour of the engagement.

\paragraph{Eye-Tracking Acquisition} 
\label{sec:eye tracking acquisition}
\noindent Eye movements were recorded with an infrared video-based eye tracker (EyeLink 1000 Plus, SR Research) at a sampling rate of 500 Hz and an instrumental spatial resolution of 0.01°.  The eye-tracker was calibrated with a 9-point grid at the beginning of the session and re-validated before each block of trials.
Participants were instructed to keep their gaze on a given point until it disappeared. If the average error of all points (calibration vs. validation) was below 1° of visual angle, the positions were accepted. Otherwise, the calibration was redone until this threshold was reached.

\paragraph{EEG Acquisition}
\noindent We recorded the high-density EEG data at a sampling rate of 500 Hz with a bandpass of 0.1 to 100 Hz, using a 128-channel EEG Geodesic Hydrocel system. A signal snippet can be seen in Figure \ref{fig:signal snippet}. The \textit{Cz} electrode served as a recording reference. The impedance of each electrode was checked before recording and was kept below 40 $k\Omega$. Additionally, electrode impedance levels were checked approximately every 30 minutes and reduced if necessary.

\subsection{Stimuli \& Experimental Design}
\label{sec:stimuli-and-experimental-design}

\begin{figure*}[t!]
    \centering
    \includegraphics[scale=0.33]{ 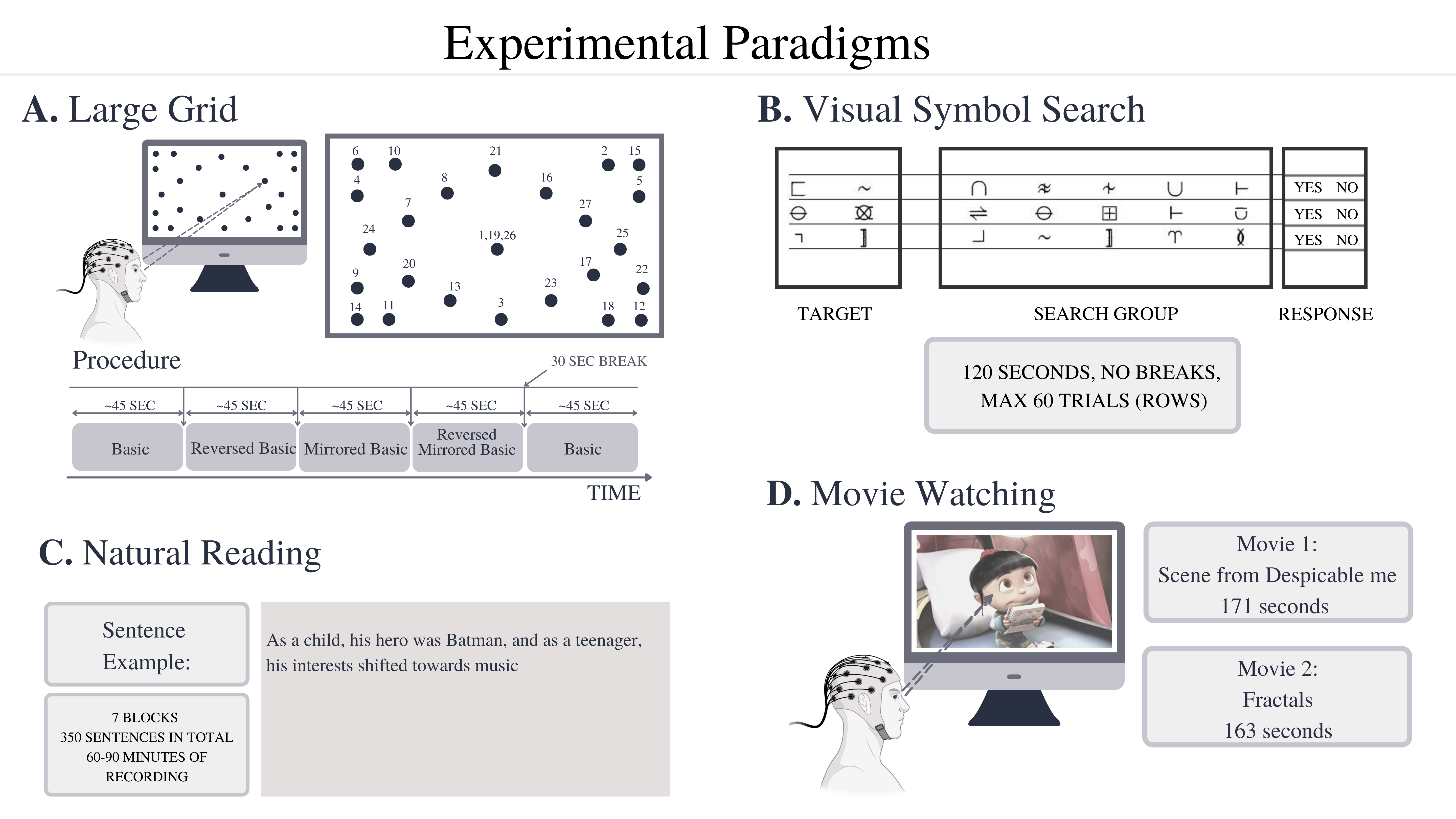}
    \caption{Schematic overview of all experimental paradigms used in our evaluation: Movie Watching, Natural Reading, Visual Symbol Search, and Large Grid.}
    \label{fig:experimental paradigm panel}
    %\vspace{-0.4cm}
\end{figure*}

\iffalse
\begin{figure}[ht!]
     \centering
     \begin{subfigure}[b]{0.5\textwidth}
         \centering
         \includegraphics[scale=0.48]{ images/movie-final.png}
         %\caption{Schematic overview of the Movie paradigm.}
         \label{fig:movie-paradigm}
     \end{subfigure}
     \hfill
     \begin{subfigure}[b]{0.5\textwidth}
         \centering
        \includegraphics[scale=0.48]{ images/reading-final.png}
         %\caption{Schematic overview of the Natural Reading paradigm.}
         \label{fig:reading-paradigm}
     \end{subfigure}
     
     \begin{subfigure}[b]{0.5\textwidth}
         \centering
         \includegraphics[scale=0.48]{ images/vss-final.png}
         %\caption{Schematic overview of the Visual Symbol Search paradigm.}
         \label{fig:vss-paradigm}
     \end{subfigure}
     \hfill
     \begin{subfigure}[b]{0.5\textwidth}
         \centering
         \includegraphics[scale=0.48]{ images/large-grid-final.png}
         %\caption{Schematic overview of the Large Grid paradigm.}
         \label{fig:large-grid-paradigm}
     \end{subfigure}
        \caption{Schematic overview of all experimental paradigms used in our evaluation: Movie Watching, Natural Reading, Visual Symbol Search, and Large Grid.}
        \label{fig:schematic-paradigms}
\end{figure}
\fi

\subparagraph{Large Grid}
We used the \textit{Large Grid} paradigm as proposed in \cite{DBLP:journals/corr/abs-2111-05100}, where participants fixate on a series of sequentially presented dots, each at one of 25 unique positions (presentation duration = 1.5-1.8s). This experimental paradigm is similar to the calibration procedures, which are used in infrared video- based eye-tracking approaches. The positions cover all corners of the screen as well as the center (see Figure \ref{fig:experimental paradigm panel}). 
Unlike the others, the dot at the center of the screen appears three times, resulting in 27 trials (displayed dots) per block.  
%5 blocks for each subject (~45 seconds/block), which equals 150 calibration points)
To record a larger number of trials and reduce the predictability of the subsequent positions in the primary sequence of the stimulus, we used different pseudo-randomized orderings of the dots presentation, distributed in five experimental blocks, as shown in Figure 3. The entire procedure was repeated 6 times during the measurement, resulting in 810 stimuli presentations per each participant.

\iffalse 
\begin{figure}[h!]
    \centering
    \includegraphics[scale=0.55]{ images/large-grid-final.png}
    \caption{Schematic view of the \textbf{Large Grid} Experimental paradigm. All dots are sequentially displayed as indicated by their number. Five blocks of dots of 45 seconds length constitute one trial. The trial is repeated 6 times for each participant.}
    \label{fig:large-grid}
\end{figure}
\fi

\subparagraph{Visual Symbol Search}
\textit{Visual Symbol Search (VSS)} paradigm is a computerized version of a clinical assessment to measure processing speed (Symbol Search Subtest of the Wechsler Intelligence Scale for Children IV (WISC-IV)~\cite{wechsler1949wechsler}) and the Wechsler Adult Intelligence Scale (WAIS-III) ~\cite{wechsler1955wechsler}). Participants are shown 15 rows at a time, where each row consists of two target symbols, five search symbols and two additional symbols that contain respectively the words “YES” and “NO”. For each row, participants need to indicate by clicking with the mouse button on the “YES” or “NO” symbol, whether or not one of the two target symbols appears among the five search symbols \cite{kastrati2021eegeyenet}. A schematic overview can be found in Figure \ref{fig:experimental paradigm panel}. For detailed event information of the dataset we refer to Appendix \ref{app:vss}.

\iffalse 
\begin{figure}[h!]
    \centering
    \includegraphics[scale=0.55]{ images/vss-final.png}
    \caption{Schematic view of the \textbf{Visual Symbol Search} Experimental paradigm. Participants work row by row and compare the two symbols from the \textit{target group} with the five symbols from the \textit{search group}. They enter their \textit{YES/NO} response after they made up their decision. The task is performed under time pressure.}
    \label{fig:vss}
\end{figure}
\fi

\subparagraph{Natural Reading}
In order further explore performance under natural viewing conditions, we test our model suite on the publicly available \textit{ZuCo 2.0} dataset \cite{zuco2}. \textit{ZuCo 2.0} is a dataset containing simultaneous eye-tracking and electroencephalography data collected during natural reading. The dataset we use contains gaze and brain activity data from nine participants reading English language sentences, both in a normal reading paradigm and in a task-specific paradigm. In the latter, participants actively search for a semantic relation type in the given sentences as a linguistic annotation task. A schematic overview of the experimental paradigm can be found in Figure \ref{fig:experimental paradigm panel}. For detailed event information we refer to Appendix \ref{app:natural-reading}.

\iffalse 
\begin{figure}[h!]
    \centering
    \includegraphics[scale=0.55]{ images/reading-final.png}
    \caption{Schematic view of the \textbf{Natural Reading} Experimental paradigm. Participants read English language sentences. Here, the normal reading paradigm is displayed.}
    \label{fig:natural-reading}
\end{figure}
\fi

\subparagraph{Movie Watching}
\label{sec:movie watching}
The \textit{Movie Watching} paradigm enables the collection of EEG data in a naturalistic viewing setting. Data were obtained during two short and highly engaging movies scenes (‘Despicable Me’ [171 seconds clip, MPEG-4 movie, the bedtime ("Three Little Kittens") scene] and ‘Fun Fractals’ [163 seconds clip, MPEG-4 movie]). A schematic view of the experimental paradigm can be found in Figure \ref{fig:experimental paradigm panel}. For detailed event information we refer to Appendix \ref{app:movie}.

\iffalse 
\begin{figure}[t]
    \centering
    \includegraphics[scale=0.55]{ images/movie-final.png}
    \caption{Schematic view of the \textbf{Movie Watching} Experimental paradigm. Participants are shown two different Movies on the screen.}
    \label{fig:movie-watching}
\end{figure}
\fi

\subsection{Preprocessing and Data Annotation}
\label{sec:preprocessing}

\paragraph{Eye-Tracking Preprocessing}
Existing literature studying eye movement generally distinguishes between three different ocular events  \cite{toivanen2015probabilistic}: saccades, fixations, and blinks. \textit{Saccades} are rapid, ballistic eye movements that instantly change the gaze position. Saccade onsets are detected using the eye-tracking software default settings: acceleration larger than 8000°/s\textsuperscript{2}, a velocity above 30°/s, and a deflection above 0.1°. \textit{Fixations} are defined as time periods while maintaining of the gaze on a single location \cite{engbert2003microsaccades}, and \textit{blinks} are considered a special case of fixation, where the pupil diameter is zero. One example of such labeled data is shown in Figure \ref{fig:signal snippet}.

%and an instrument spatial resolution of less than 0.01$\degree$ root mean square (RMS) of the distances between successive samples \cite{eegeyenet}. 
%The ET was calibrated with a 9-point grid before each recording. 
%In a validation step, the ET calibration was repeated until the error between two measurements at any point was less than 0.5$\degree$, or the average error for all points was less than 1$\degree$.
%Participants were seated at a distance of 68 cm from a 24-inch monitor with a resolution of $800 \times 600$ pixels.
%Unlike the others, the dot at the center of the screen appears three times, resulting in 27 trials (displayed dots) per block, each dot is displayed for $1.5$ to $1.8$ seconds. 
%The positions of the dots were selected to ensure coverage of all corners of the screen as well as the center (see Figure~\ref{fig:large-grid}). The shape of the grid and its use for eye gaze estimation follows the work from \cite{son}. 
%Given that \cite{son} used the Large Grid paradigm for functional Magnetic Resonance Imaging (fMRI), the length of the stimulus was adapted and the number of repetitions to the setup described in \cite{eegeyenet}. To record a larger number of trials and reduce the predictability of the subsequent positions in the primary sequence of the stimulus, different pseudo-randomized orderings of the dots presentation are used, distributed in five experimental blocks, as shown in Figure~\ref{fig:large-grid}.
\paragraph{Electroencephalography Preprocessing}
EEG data is often contaminated by artifacts produced by environmental factors, e.g., temperature, air humidity, as well as other sources of electromagnetic noise, such as line noise~\cite{kappenman2010effects} and therefore requires preprocessing in the form of artifact cleaning or artifact correction \cite{keil2014committee}.
Our EEG preprocessing included detecting and interpolating bad electrodes and filtering the data with a 0.1 Hz high-pass filter. With this ``minimal'' preprocessing pipeline,  the preprocessed data still retains ocular activity, which can be used to infer the ocular events (the signal of interest in the present study). The detailed preprocessing pipeline is described in Appendix \ref{appsec-eeg}.
%The EEG preprocessing was conducted with the open-source MATLAB toolbox pipeline Automagic \cite{pedroni2019automagic}, which combines state-of-the-art EEG preprocessing tools into a standardized and automated pipeline. The EEG preprocessing consisted of the following steps: First, bad channels were detected by the algorithms implemented in the EEGlab plugin \texttt{clean\_rawdata}.\footnote{\url{http://sccn.ucsd.edu/wiki/Plugin\_list\_process}} Detected bad channels were automatically removed and later interpolated using a spherical spline interpolation. Subsequently, residual bad channels were excluded if their standard deviation exceeded a threshold of $25 \mu V $. Very high transient artifacts ($> \pm100 \mu V$) were excluded from calculating the standard deviation of each channel. Next, line noise artifacts were removed by applying Zapline \citep{de2020zapline}. However, if this resulted in a significant loss of channel data ($>$ 50\%), the channel was removed from the data.  
%Due to the poor data quality and missing parts of the data, we removed 14 participants' recordings from the dataset. Therefore, the final sample used for experiments consists of recordings from 168 subjects.

\paragraph{EEG \& Eye-Tracking Synchronization}
In the next step, the EEG and eye-tracking data were synchronized using the EYE-EEG toolbox \citep{dimigen2011coregistration} to enable EEG analyses time-locked to the onsets of fixations and saccades, and subsequently segment the EEG data based on the eye-tracking measures.
The synchronization algorithm first identified the ``shared'' events between eye-tracking and electroencephalography recordings (start end trigger of each experimental paradigm). Next, a linear function was fitted to the shared event latencies to refine the start- and end-event latency estimation in the eye tracker recording. Finally, the synchronization quality was ensured by comparing the trigger latencies recorded in the EEG and eye-tracker data. 
As a result, all synchronization errors did not exceed 2 ms (i.e., one data point).

\section{EEG Segmentation Task \& Metric}
\begin{figure}[t]
    \centering
    \includegraphics[width=0.5\textwidth]{ 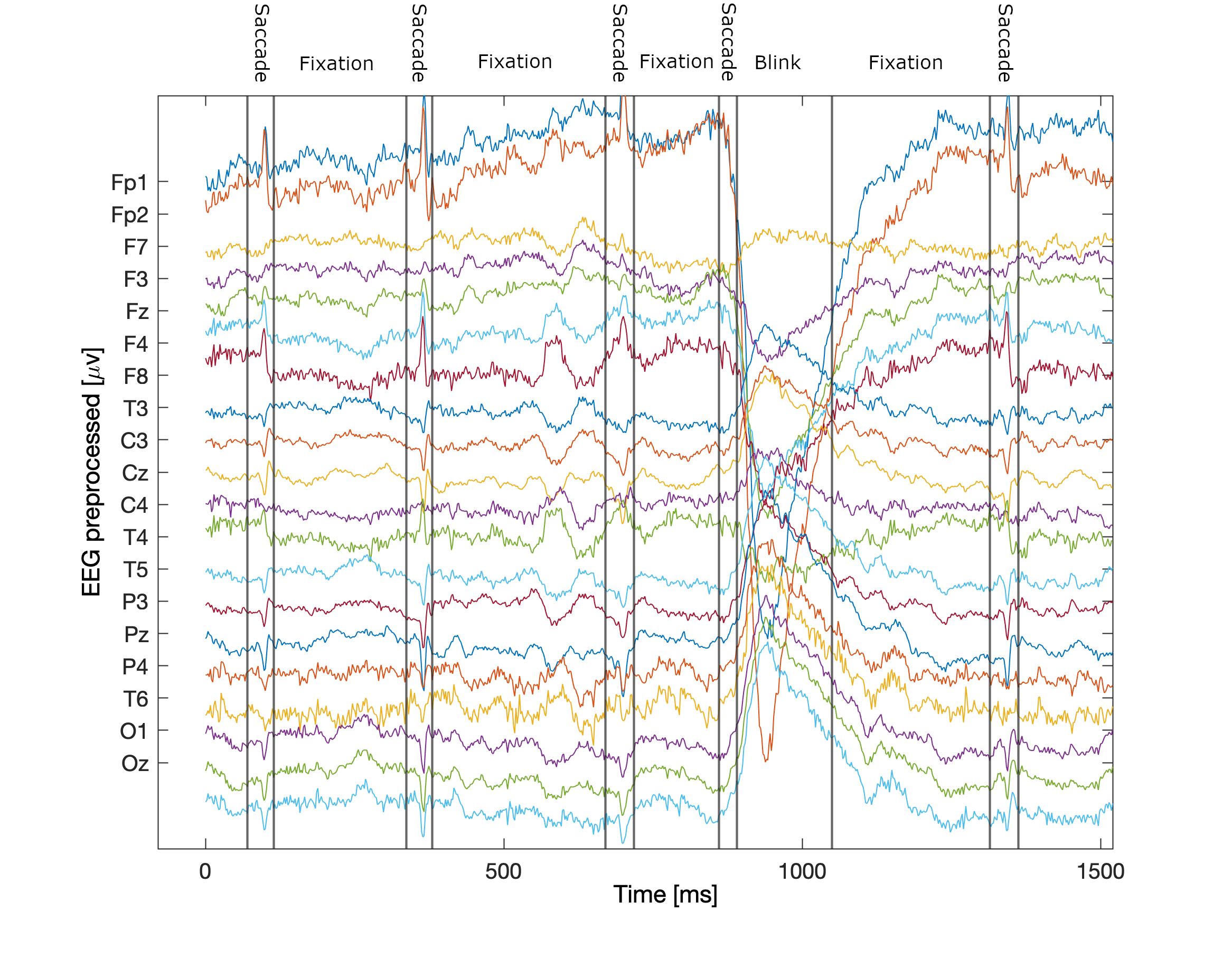}
    \caption{EEG data and the segments with ocular events.}
    \label{fig:signal snippet}
    \vspace{-0.8cm}
\end{figure}

%Based on the collected data from 72 participants, 
For each of the four datasets we approached the task of segmenting the EEG data into three different ocular events: fixations, saccades and blinks. As illustrated in Figure \ref{fig:signal snippet}, the goal of this task was to classify each time point (i.e 2ms) of the 128-channel EEG stream of data as a unique event.

\paragraph{Data Structure}
\label{sec:DETR_dataprep}
Since our model operates on samples of arbitrary length (e.g. 500 time steps), we cut fixed size samples from a participants' continuous data stream. Each participant was randomly assigned to either training, validation or test set.  We assigned 70\% of the participants to our training data set, and 15 \% to validation and test set, respectively. Details on the number of events in each dataset are summarized in Table \ref{tab:all_datasets_event_distibution}. %According to our policy of feeding sequence lengths of size 500 to our model, we end up with 68179 samples in the training set, and 15174 as well as 15468 samples in validation and test set.
The prepared dataset is publicly available and can be found online in an OSF repository \footnote{\href{https://osf.io/dr6zb/}{https://osf.io/dr6zb/}}.

\paragraph{Evaluation Metric}

Since our dataset is heavily imbalanced, we used F1-score as an evaluation metric. This evaluation procedure was also proposed in comparative literature such as in \cite{perslev2019utime}. 
Moreover, the F1 score is often used for semantic segmentation in the computer vision literature (among other similar metrics such as Intersection-over-Union), where segmentation is achieved by classifying single, atomic pixels/time steps \cite{panoptic_segmentation}. Depending on the setting, a background class can be introduced.
For the purpose of our study, we opted to segment the complete time series, where each time point (i.e. 2ms) was classified as either belonging to a blink, a fixation or a saccade. 
%Therefore, the events did not overlap, and there were no regions in the signal that can be considered as background. 

% It's noteworthy that this corresponds to the semantic segmentation definition found in computer vision literature \cite{panoptic_segmentation}.

%Several other evaluation metrics are also used for segmentation. %In addition, in computer vision literature, one distinguishes between different types of segmentation tasks: semantic, instance, and panoptic segmentation \cite{TODO:panoptic segmentation paper}, each task with varying metrics of evaluation.

\begin{table}[t]
    \centering
    \scalebox{0.68}{
    \begin{tabular}{lcccr}
         \toprule
        {Dataset}  & \specialcell{\# Fixations \\ Fixation Time} & \specialcell{\# Saccades \\ Saccade Time} & \specialcell{\# Blinks \\ Blink Time}  & \specialcell{\# Total Events \\ Total Time} \\
        \midrule
        Large Grid & \specialcell{108416 \\ 1522 min} & \specialcell{108500 \\ 109 min} & \specialcell{10187 \\ 19 min} & \specialcell{227103 \\ 1650 min} \\
        \midrule
        {VSS} & \specialcell{42383 \\ 142 min} & \specialcell{42375 \\ 32 min} & \specialcell{948 \\ 1 min} & \specialcell{85706 \\ 175 min} \\
        \midrule
        {Natural Reading} & \specialcell{72251 \\ 263 min} & \specialcell{72236 \\ 58 min}  & \specialcell{5301 \\ 10 min} & \specialcell{149788 \\ 332 min } \\ 
        \midrule
        Movies & \specialcell{30399 \\ 220 min} & \specialcell{30387 \\ 27 min}  & \specialcell{3559 \\ 7 min} & \specialcell{64345 \\ 254 min} \\ 
        \bottomrule
    \end{tabular}
    }
    \caption{Event distribution of all ocular event datasets. We report the number of events and the total event length of all three classes. The event frequency depends strongly depends on the experimental paradigm. Detailed information about each dataset can be found in Appendix \ref{app:event details}.}
    \label{tab:all_datasets_event_distibution}
    %\vspace{-0.6cm}
\end{table}

%Our ground truth signal provides annotations in the form of start and end time of each event, as well as the start and end position of saccades and the average position of fixations. 

%\begin{figure}[t]
%    \centering
%    \includegraphics[width=0.45\textwidth]{images/eeg_prediction.png}
%    %\vspace{0.3cm} 
%    
%    \includegraphics[width=0.45\textwidth]{images/eeg_true.png}
%    \caption{EEG Data Example where the top image is the prediction (DETR) and the bottom image is the true image. Background %(unlabeled) are saccades.}
%    \label{fig:eeg_data_sample}
%    %\vspace{-0.3cm}
%\end{figure}

\subsection{Potential Issues and Challenges}
\label{sec:potential-issues}

\paragraph{Micro Events}
\label{sec:microevents}
In certain experimental paradigms, such as Large Grid, participants tend to correct their gaze before fixating on the final location (e.g. a participant moves the eye gaze to the consecutively presented dots). This leads to phases of very short saccades and fixations. The alternating and brief phases of fixations and saccades make it hard for any convolution-based classification model to accurately classify events based on their immediate neighborhood. An additional restrictive factor is the limited temporal resolution of the infrared video-based ground truth signal (see Section \ref{sec:eye tracking acquisition}). 
%illustrates another problems for detection models. Since the micro saccades split larger fixations into smaller ones, the model is unsure about the number of events in a given time-frame.

%\begin{figure}[t]
%    \centering
%    \includegraphics[width=0.45\textwidth]{images/micro_sacc_pred.png}
%    %\vspace{0.3cm} 
%    
%    \includegraphics[width=0.45\textwidth]{images/micro_sacc_true.png}
%    \caption{EEG Data Example with short saccades. The model has difficulty to properly find all the saccades in the sample. Top is prediction (DETR) and bottom is ground-truth.}
%    \label{fig:eeg_microsacc_sample}
%    %\vspace{-0.3cm}
%\end{figure}

\paragraph{Ambiguity in the Signal}
\label{sec:ambiguity-signal}
%We can coarsely differentiate between vertical and horizontal saccades by imagining 90 degree areas for each of the classes on positive as well as negative axis. 
From a signal processing perspective, vertical saccades are hard to differentiate from blinks \cite{kleifges2017blinker}. This may lead to a higher number of wrong predictions in the respective classes. 

\paragraph{Dataset Imbalance}
\label{sec:dataset-imbalance}
Fixation events last, on average, much longer than saccades or blinks (see Appendix \ref{app:event details} for a detailed event analysis of all experimental paradigms). While fixations and saccades are balanced in event occurrence, each experimental paradigm exhibits a significant temporal imbalance of ocular events through the nature of the eye gaze itself. To counter this temporal imbalance, we introduce \textit{biased sampling} to the training procedure of our baseline model suite and our DETRtime architecture. This strategy assigns each data sample a sampling probability based on the event types. In our case, higher weight is given to samples containing more saccade or blink events, thus introducing a bias. The data loader then draws samples into batches based on this distribution. Therefore, model parameters get updated more often on samples containing minorities. 
%\paragraph{Signal Noise}
%\label{sec:signal-noise}. DONT USE THIS SECTION!!!!!!!!
%The human brain should not be capable of producing a voltage of more than $100 \mu V$ measured at EEG electrodes. Therefore, $100 \mu V$  can be seen as an upper bound of valid measurement values. However, in some samples, we experience failing electrodes that exceed this value, producing disliked signal noise. Based on this prior knowledge, we decided to cut the continuous stream into multiple parts if the mean absolute amplitude value across all EEG electrodes exceeds a threshold of 150 $\mu$V. In practice, this policy is sufficiently liberal in allowing sampling large substreams from participants' continuous streams in a majority of cases. However, in minor cases and the presence of an increased amount of noise in the data, the strategy above leads to a slight stream fragmentation into up to 5 \textit{clean} substreams that fulfil physiologically reasonable conditions. 
%As can be seen in figure \ref{fig:eeg_noise_sample}, some samples contain measurement errors where the measured voltage exceeds reasonable values.

%\paragraph{Imbalanced Data}
%\label{sec:signal-noise}
%??
%As can be seen in figure \ref{fig:eeg_noise_sample}, some samples contain measurement errors where the measured voltage exceeds reasonable values.

\section{Baselines}
%In this section we establish baselines for the task and introduce our two main approaches, namely classification and object detection. 

\label{sec:baselines}
%\subsection{Establishing Baselines}
%\label{sec:baselines}
We run extensive experiments on the proposed task in order to provide baselines of different complexity for the EEG segmentation task. As part of our baseline model suite, we compared models based on classical machine learning techniques (see Appendix \ref{app:standard machine learning models detailed}) to established deep neural network architectures. In addition to that, we included recently proposed models that are tailored explicitly for EEG data \cite{eegnet,perslev2019utime}. Detailed descriptions of all Deep Learning-based baseline models can be found in Appendix \ref{app:deep learning baselines}. The state-of-the-art time series segmentation models U-Time and SalientSleepnet are covered in Appendix \ref{app: state-of-the-art-baselines}.

\section{DETRtime}
\label{sec:detr}

End-to-End Object DEtection with TRansformers (DETR) \cite{detr} was originally proposed for object detection in natural images. 
%In addition, it is relatively straightforward to extend the original model to also predict segmentation masks. 
%In this paper, we follow the original implementation of DETR with a few changes and adapt it for time-series. 
We argue that the object detection objective as defined in \cite{detr} can be adapted to predict discrete time-segments, yielding several benefits over the classical semantic segmentation approach. In addition, it is relatively straightforward to process predicted regions to one single segmented time series.
We introduce the DETRtime framework as a time-series adaptation of the the original DETR architecture. The code is publicly available 
\footnote{\href{https://github.com/lu-wo/DETRtime}{https://github.com/lu-wo/DETRtime}}.

\subsection{Architecture}

\begin{figure*}
    \centering
    \includegraphics[width=1.0\textwidth]{ 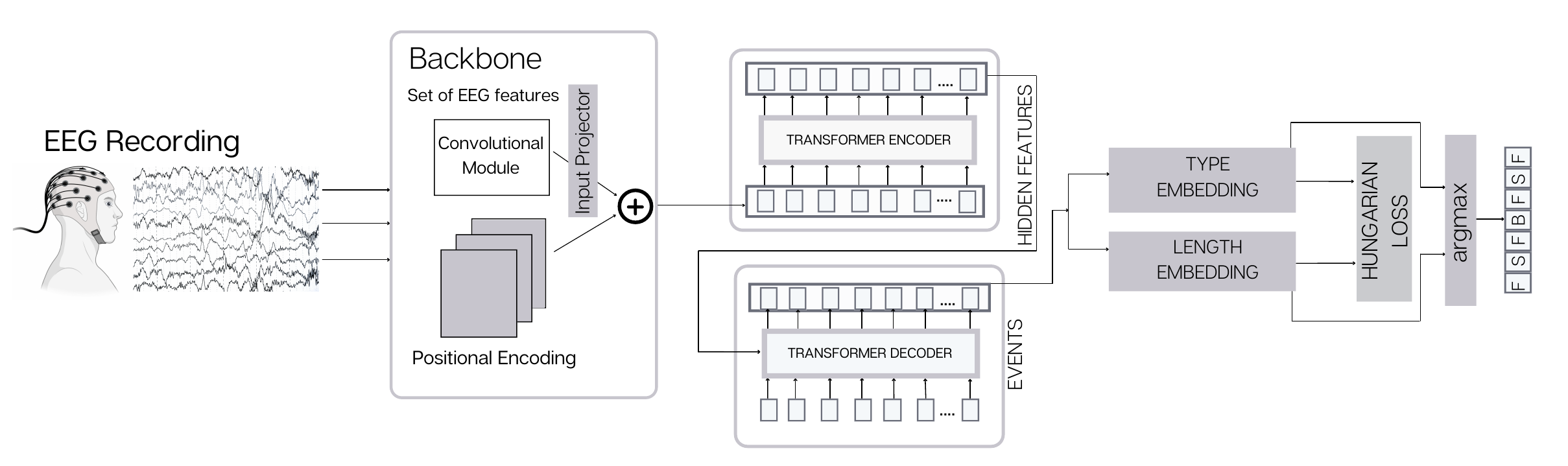}

    \caption{DETRtime Architecture is composed of a backbone model, a positional embedding followed by a transformer and finally feed-forward networks that map the hidden features to the events and their length.}
    \label{fig:DETR_architecture} 
    %\vspace{-0.4cm}
\end{figure*}

The architecture of the model is illustrated in Figure \ref{fig:DETR_architecture}. At a high level, it consists of a backbone, transformer, and an embedding layer (for type of event, and the length and position of the event). In the following section, we explain each building block of this architecture.
%.  

\paragraph{Backbone}
First, a convolutional backbone learns a representation of the input signal. In the original DETR implementation, the backbone model is a \textit{Resnet50} model pretrained on images. For our case, we experimented with several convolutional models: classical CNN, Pyramidal CNN, InceptionTime, and Xception, all part of our baseline model suite (see Section \ref{sec:baselines}). The backbone module is composed of 6 layers of convolutional models with  skip connections between layers $i$ and $(i + 2) \% 3$ and a final projection layer that maps the input to a shape suitable for the transformer.

\paragraph{Positional Embedding}
As usual for transformers, the features of the backbone are supplemented with a positional encoding before being passed into a transformer encoder since we must inject some information about the relative or absolute position of the sequence embeddings. We use sine functions for positional encoding as described in \cite{attentionIsAllYouNeed}.

\paragraph{Transformer}
The transformer consists of an encoder and decoder. The transformer encoder learns a representation of the sequence, which is fed together with the learned queries into the decoder. The transformer decoder then processes a small but fixed number of  $N$ learned \textit{event queries}, and additionally attends to the encoder output. The output of the decoder produces encoded features for these $N$ event queries. Therefore, we obtain an upper bound of $N$ detectable events in a given sample. For our purposes, $N=20$ is sufficient for almost all EEG data streams with a length of 1 second. The output of the decoder is then used to predict the properties of each event. 

\paragraph{Event Prediction} In order to predict events, we make use of a 3-layer perceptron to map the encoded features of the $N$ event queries to their respective type, position and length. These $N$ predictions represent the events that are later used to produce the final segmentation.

\subsection{Object Detection as Time Region Prediction}
As a single forward pass of the network up to the transformer decoder produces N unordered predictions, one of the main challenges during training is defining a well-suited loss function to evaluate how well these N predictions detect the events contained in the ground truth signal. The Hungarian loss proposed in DETR \cite{detr} solves this problem by matching each predicted object to one ground-truth object and then minimizing the objective
\begin{align*}
 \ell =  \ell_{L_1} + \ell_{GIoU} + \ell_{CE}
\end{align*}
where $\ell_{L_1}$ and $\ell_{GIoU}$ is the L1 distance between the boundaries and the general intersection over the union of the target and predicted region, respectively. Combined, both measures optimize the fit between the target bounding box and the predicted bounding box of the respective objects. $\ell_{CE}$ is the Cross-Entropy loss on the predicted labels of each region, optimizing the class prediction. Unmatched predictions are expected to predict an additional ``no class`` dummy class, added as a slightly penalized class to the classification loss.  
This object detection objective translates well into time series segmentation, as time segments are perfectly localized by their bounding boxes. Thus segmentation is equivalent to finding the bounding boxes.\\
By restricting our prediction to N discrete regions, we regularize the region prediction to an extent that does not exist in classical semantic segmentation. Namely, the prediction is limited to N smooth regions and cannot easily produce artifacts. We limit the class imbalance similar to semantic segmentation-based methods. Furthermore, since classification is only considered on discrete regions, the temporal class imbalance encountered in other models is not an issue when training DETRtime. 

%\paragraph{Architecture Changes}
%\label{sec:DETR_architecture}
%Since we give up one dimension compared to natural images, we adapted the DETR loss such that it ignores the y-coordinates of our predictions and replace the \textit{Resnet50} by a custom 1D convolutional backbone which is based on the \textit{Pyramidal CNN} architecture introduced in \ref{sec:pyramidal-cnn}.

\subsection{Optimization}
\label{sec:optimization}
In our training procedure, we used the Adam optimizer with a learning rate $\eta = 1e$-4. To further counter class imbalance, we selected batches of size B = 32 on the fly and used biased sampling, as introduced in Section \ref{sec:dataset-imbalance}. For instance, we experimented with feeding our DETRtime model an elevated number of samples containing minority events (saccades and blinks). In our best-performing model (see Table \ref{tab:results}), only samples containing blinks were given an increased probability of being chosen by the dataloader.

%Note that our sampling strategy is dynamic and our framework allows reassigning these probabilities on resuming model training.
%Thus, this strategy allows precise refinement of our models weaknesses 
%(TODO LUKAS).

\subsection{Inference}
The immediate output of the model is not a full segmentation but instead N discrete, possibly overlapping time segments with different class confidences. We processed this into a segmentation prediction according to the following heuristics:

\begin{itemize}
    \item We ignored all segments predicting the dummy class with the highest confidence.
    \item Each time step was assigned to the highest confidence class of an event overlapping it.
    \item Since predicted segments were allowed to float freely, it was not ensured that each time step was covered by a prediction. Time steps without a predicted event were assigned to the majority class.
\end{itemize}

This heuristic can be applied to produce a time series of arbitrary temporal resolution. In order to appropriately compare the above segmentation to classical semantic segmentation models and methods using standard metrics such as the F1 score, we applied the above heuristic to transform the prediction constituted of discrete segments into a time series. \\
% Note that in the time series case, a instance segment as predicted by its borders is already a perfect segmentation. This is not the case for the usual image case, necessitating additional steps for panoptic segmentation.
% Note that in image segmentation, for object detection, having bounded square boxes around the object is not enough for segmenting the object. Thus, an additional segmentation module is typically used on top of the detection module. 
% However, in our case, we have 1D signals, and bounding boxes can be used for segmenting the signal. 
We found that the proposed heuristic performs well, and using a separate segmentation pipeline on the transformer output does not bring any advantages (see Appendix \ref{app:segmentation pipeline}). In Table \ref{tab:architecture_performance}, we report the size of each model and its GPU inference time. Finally, a detailed explanation of all hyperparameters of the DETRtime architecture and the optimization procedure are reported in Appendix \ref{app:hyperparamsDETR}.

\begin{table}[h]
    \centering
    \scalebox{0.85}{
    \begin{tabular}{l|ccr}
        Model & \# Parameters & {Size} & Inference Time\\
        \toprule
        CNN & 790K & {3.01MB} & 3.76ms \\
        
        Pyramidal CNN & \specialcell{ 723K}& { 2.76MB} &  3.52ms \\
        
        EEGNet & \specialcell{35K } & {0.13MB} & 1.49ms \\
        
        InceptionTime & \specialcell{939K } & { 3.59MB} & 11.17ms \\
        
        Xception & \specialcell{1277K } & { 4.89MB} & 8.94ms \\
        
        LSTM & 912K & 3.65MB & 33.60ms \\
        
        biLSTM & 2098K & 8.40MB & 67.27ms \\
        
        CNN-LSTM & 30433K & 121.80MB & 43.74ms \\
        
        \midrule
        SalientSleepNet & 187453K & 750.35MB & 30.81ms \\
        U-Time & 67966K & 259.27MB & 10.69ms \\ 
        DETRtime & 7725K & 29.49MB & 15.42ms \\ 
        \bottomrule
    \end{tabular}
    }
    \caption{Complexity and inference time of our model suite. We provide the number of parameters of each model as well as the file size in Megabyte. Inference was performed on a single GPU (Nvidia Titan Xp) and is given in milliseconds. We group the models into established deep learning models (above) as well as specialized segmentation models (below).}
    \label{tab:architecture_performance}
    %\vspace{-0.4cm}
\end{table}

\section{Results}
\label{sec:eval-results}

\begin{table*}[h]
    \scalebox{0.87}{
    \begin{tabular}{lcccc|cccc|cccc|cccc|}
    \toprule
    &    \multicolumn{4}{c}{Large Grid}
    &   \multicolumn{4}{c}{VSS} 
          &  \multicolumn{4}{c}{Reading}  
      &  \multicolumn{4}{c}{Movies}   
      
     \\
    \cmidrule(r){2-5}
    \cmidrule(r){6-9}
    \cmidrule(r){10-13}
    \cmidrule(r){14-17}
   Model & fix & sac & blk & avg & fix & sac & blk & avg  & fix & sac & blk & avg & fix & sac & blk & avg
     \\
     \midrule
     % CNN LSTM & 0.85 & 0.31 & 0.58 & 0.58 & 0.86 & 0.54 & 0.71 & 0.71 & 0.75 & 0.44 & 0.36 & 0.52 & 0.91 & 0.00 & 0.66 & 0.00 & 0.00 & 0.31 \\
    U.a.r. & 0.49 & 0.12 & 0.02 & 0.21 & 0.47 & 0.24 & 0.01 & 0.24 & 0.47 & 0.23 & 0.05 & 0.25  & 0.48 & 0.16 & 0.05 & 0.23 
     \\
     Prior & 0.92 & 0.08 & 0.01 & 0.34 & 0.81 & 0.18 & 0.01 & 0.33 & 0.79 & 0.18 & 0.03 & 0.33 & 0.86 & 0.11 & 0.03 & 0.33 
     \\
     Most Freq. & 0.96 & 0.00 & 0.00 & 0.32 & 0.89 & 0.00 & 0.00 & 0.30 & 0.89 & 0.00 & 0.00 & 0.30  & 0.93 & 0.00 & 0.00 & 0.31  
     \\
     \midrule
     kNN & 0.60 & 0.38 & 0.77 & 0.58 & 0.89 & 0.23 & 0.23 & 0.45  & 0.88 & 0.32 & 0.12 & 0.44 & 0.94 & 0.28 & 0.18 & 0.47  
     \\
     DecisionTree & 0.55 & 0.39 & 0.66 & 0.52 & 0.71 & 0.30 & 0.00 & 0.34 & 0.71 & 0.30 & 0.00 & 0.34   & 0.78 & 0.24 & 0.00 & 0.34 
     \\
     RandomForest & 0.67 & 0.36 & 0.82 & 0.61 & 0.88 & 0.27 & 0.14 & 0.43 & 0.89 & 0.36 & 0.00 & 0.42 & 0.93 & 0.32 & 0.21 & 0.48  
     \\
     RidgeClassifier & 0.67 & 0.14 & 0.82 & 0.54 & 0.83 & 0.20 & 0.00 & 0.34  & 0.77 & 0.28 & 0.06 & 0.37 & 0.90 & 0.17 & 0.21 & 0.43 
     \\
     \midrule
     CNN & 0.99 & 0.83 & 0.58 & 0.80 & 0.97 & 0.89 & 0.29 & 0.72 & 0.88 & 0.45 & 0.33 & 0.55 & 0.96 & 0.70 & 0.55 & 0.74 
     \\
     PyramidalCNN & 0.99 & 0.83 & 0.52 & 0.78 & 0.97 & 0.86 & 0.20 & 0.68 & 0.87 & 0.44 & 0.31 & 0.54  & 0.96 & 0.71 & 0.58 & 0.75 
     \\
     EEGNet & 0.99 & 0.80 & 0.55 & 0.78 & 0.97 & 0.87 & 0.53 & 0.79  & 0.89 & 0.49 & 0.33 & 0.57 & 0.96 & 0.65 & 0.32 & 0.64 
     \\
     InceptionTime & 0.99 & 0.83 & 0.53 & 0.78 & 0.97 & 0.89 & 0.29 & 0.72 & 0.88 & 0.45 & 0.31 & 0.55 & 0.96 & 0.72 & 0.58 & 0.75 
     \\
     Xception & 0.99 & 0.84 & 0.63 & 0.82 & 0.98 & 0.90 & 0.45 & 0.77 & 0.89 & 0.43 & 0.45 & 0.59  & 0.96 & 0.70 & 0.58 & 0.75 
     \\
     LSTM & 0.98 & 0.71 & 0.49 & 0.73 & 0.97 & 0.86 & 0.18 & 0.67 & 0.86 & 0.47 & 0.29 & 0.54  & 0.94 & 0.62 & 0.49 & 0.68 
     \\
     biLSTM & 0.98 & 0.79 & 0.47 & 0.75 & 0.97 & 0.86 & 0.28 & 0.70 & 0.87 & 0.44 & 0.27 & 0.53  & 0.96 & 0.69 & 0.57 & 0.74 
     \\
     CNN-LSTM & 0.99 & 0.84 & 0.52 & 0.78 & 0.97 & 0.88 & 0.24 & 0.70 & 0.87 & 0.44 & 0.34 & 0.55  & 0.96 & 0.70 & 0.56 & 0.70 
     \\
     \midrule
     SalientSleepNet & 0.99 & 0.81 & 0.51 & 0.77 & 0.97 & 0.86 & 0.24 & 0.69 & 0.85 & 0.40 & 0.43 & 0.56  & 0.96 & 0.62 & 0.52 & 0.70  
     \\
     U-Time & 0.99 & 0.82 & 0.88 & 0.90 & 0.90 & 0.70 & 0.79 & 0.79 & 0.86 & 0.47 & 0.72 & 0.69 & 0.96 & 0.70 & 0.62 & \textbf{0.76}  
     \\
     DETRtime & 0.99 & 0.87 & 0.90 & \textbf{0.92} & 0.86 & 0.78 & 0.82 & \textbf{0.86} & 0.90 & 0.55 & 0.80 & \textbf{0.75} & 0.96 & 0.69 & 0.63 & \textbf{0.76} 
     \\
    \bottomrule 
  \end{tabular}
  }
    \caption{Results on the ocular event datasets. We report the results of the best hyperparameter configuration found for each model. Additionally, we report F1 scores of all classes as well as the (macro) average F1 score. The best results of the average F1 score are displayed in bold. Hyperparameter configurations as well as training properties can be found in Appendix \ref{app:hyperparamsBaseline}. Standard Machine Learning models (see Appendix \ref{app:standard machine learning models detailed}) are trained on a a single time step (1, 128) basis.}
    \label{tab:results}
    %\vspace{-0.1cm}
\end{table*}

In Table \ref{tab:results}, we report the results of our experiments on the four datasets introduced in Section \ref{sec:setup-data}. In addition to that, we perform experiments in applications other than eye-tracking, namely the EEG sleep stage segmentation task (see Appendix \ref{app:sleep-staging-task}). 
%The segmentation models U-time and SalientSleepNet model were originally developed with this task in mind. 
Our experiments show the generalization of our model across different experimental paradigms and tasks. %Finally, the novel dataset (Large Grid paradigm) is explicitly suitable for the segmentation of EEG data in ocular events, making possible to use the dataset for pretraining the models (Appendix \ref{app:pretraining-on-large-grid}).

\paragraph{Standard Machine Learning Models} First, we compared some of the most widely used classical machine learning models: kNN, Decision Tree, Random Forest and Ridge Classifier. 
%We can observe in Table \ref{tab:results} that the best performing model in this group is Random Forest with an average F1 score of 0.613. 
We can observe in Table \ref{tab:results} that the best performing model in this group throughout all datasets was Random Forest, achieving average F1 scores between 0.48 (Movies) and 0.61 (Large Grid). 
%As stated in the literature \cite{kleifges2017blinker} blinks are easily detected by simple techniques like thresholding on specific EEG/EOG electrodes. We can observe this also in our Table \ref{tab:results}, which shows that all standard machine learning models have the best F1 score (ranging from 0.66 - 0.82) for the blink events compared to the F1 score of the other two ocular events (saccades and fixations). Blinks are, however, difficult to distinguish from vertical saccades, resulting in a low F1 score for this ocular event. While Random Forest has an F1 score of 0.818 for the blink events, the F1 score for the saccade events is only 0.358. In particular, we can observe that all standard machine learning models struggle to detect the saccade events with a recall ranging between 0.08 and 0.411. For the fixation task, the standard machine learning models perform similar, with the exception of Ridge Classifier, which has a recall of fixation events of 0.911.
As can be seen in Table \ref{tab:results}, all standard machine learning models struggled to detect saccades and blinks on more naturalistic experimental paradigms (Movies, Reading) and on the Visual Symbol Search (VSS) dataset.  Fixations were detected fairly well (kNN achieving 0.94 F1 score on Movies). Finally, blinks were detected better on the Large Grid paradigm (Random Forest and Ridge Classifier achieving a blink F1 score of 0.82).
%On the other hand, the precision score of the Ridge Classifier is much lower (0.532). We can conclude that Ridge Classifier predicts most of the time fixation event resulting in high score in recalling/detecting fixations, however at the same time, this results in a very low score in recalling saccades (recall score 0.08). 
%Nonetheless, all standard machine learning models perform with a score in the range of 0.5-0.6, which is significantly better than our best naive baseline (predict each event by sampling with a prior distribution), which has a score of 0.33.
It is worth mentioning that on all datasets, classical machine learning models outperformed the naive baselines (e.g. prior distribution achieving 0.33 vs. Random Forest achieving 0.48 F1 score on Movies dataset).

\paragraph{Deep Learning Baselines} We compared eight deep learning-based models (CNN, Pyramidal CNN, EEGNet, InceptionTime, Xception, LSTM, biLSTM and CNN-LSTM) that are not particularly tailored for segmentation tasks but classify each time point with one of the ocular events. All deep learning architectures performed significantly better than the naive baseline and outperformed all reported standard machine learning methods. 
%Remarkably, all deep learning-based models have a similar or worse average F1 score than the standard machine learning models. In comparison to the standard machine learning models, deep learning models take a sequence of EEG data and only then predict a class for each time point. On the other hand, using the standard machine learning techniques, we predict the class of each time point separately. We can observe that the deep learning models have a higher F1 score for fixation and saccade compared to the standard machine learning ones; however, they perform much worse in the blink section. The best performing model is GazeNet, with an F1 score of 0.6. We can also observe that all deep learning models have very high precision for the fixation events, resulting in a much better F1 score for this column, as compared to the F1 score of standard machine learning models. In particular, GazeNet achieved the best F1 score (0.946) in the fixation task. GazeNet is also one of the best performing models in the saccade task with an F1 score of 0.625, together with InceptionTime, which outperformed all aforementioned deep learning models, with an F1 score of 0.642. As mentioned earlier, the deep learning models perform remarkably worse in the blink task, where the F1 score of all deep learning models ranges between 0.03 - 0.11. 
On the Large Grid paradigm, Xception achieved an average F1 score of 0.82, outperforming all other baselines in its group. Conversely, on the Visual Symbol Search paradigm, the best performing model was EEGNet, specifically designed to process EEG data, achieving an average F1 score of 0.79.
All deep learning baselines performed significantly worse on the Natural Reading paradigm, with the best performing Xception achieving an average F1 score of 0.59.
A similar image is drawn on the Movies dataset, where all deep learning baselines performed similarly well, with Xception and InceptionTime achieving average F1 scores of 0.75. These results suggest that the Natural Reading paradigm data is harder to segment than Movies. While the Movies paradigm is the most naturalistic evaluated paradigm, the Reading task mainly contains eye movements from left to right. Throughout all four datasets, the deep learning baselines achieve very high F1 scores for the fixation class. Nevertheless,  saccades and blinks are harder to detect, which stems from the ambiguous nature of their signals. In addition, blinks are challenging to distinguish from vertical saccades (see Section \ref{sec:ambiguity-signal}), resulting in lower F1 scores for both classes.

 %It is to note that within the group of deep learning baselines, both convolution-rbased models as well as recurrent models such as the LSTM architectures are able to compete with each other. The bidirectional LSTM achieved an average F1 score of 0.74 on the Movies paradigm, only surpassed by Xception and InceptionTime models. 

\paragraph{SalientSleepNet, U-Time \& DETRtime} 
SalientSleepNet and U-Time models were explicitly designed for the segmentation of psychophysiological datasets.  As demonstrated in Table \ref{tab:architecture_performance}, U-Time and DETRtime, outperformed all the other models (including SalientSleepNet) by a considerable margin. In addition,  SalientSleepNet performs well, achieving F1 scores significantly better than standard machine learning algorithms and naive baselines. Nevertheless, on average, it does not outperform other deep learning models. The gap between the two leading models and the other deep learning-based models is the largest in the Large Grid paradigm. U-Time surpasses the basic CNN architecture by 0.12, achieving an average F1 score of 0.90. This result is only reached by DETRtime, achieving an average F1 score of 0.92. On Reading and VSS paradigms, DETRtime outperforms the runner-up U-Time by a large margin, namely 0.75 vs 0.69 average F1 score on the Reading paradigm and 0.86 vs 0.79 F1 score on the VSS paradigm. On the Movies paradigm, both U-Time and DETRtime achieve an average F1 score of 0.76. Finally, it is worth noting that both U-Time and DETRtime consistently achieve high F1 scores on all datasets. %Other than the rest of the Deep Learning-based models, U-Time and DETRtime detect blinks better on Reading, VSS and Large Grid datasets, even surpassing their performance on the saccade class. 

\section{Discussion}
\label{sec:discussion}
%. add , that the model does not just learn from the brain activity itself, but that the measured EEG signal includes brain activity and ocular activity. 

In this paper, we introduced DETRtime, a novel time-series segmentation approach inspired by the recent success of the DETR  architecture \cite{carion2020end}. Furthermore, we showed that deep learning models designed initially for image segmentation are also well-suited for EEG time-series data. 

Here, we decomposed the continuous data stream into three types of ocular events: fixations, saccades, and blinks. Additionally, we presented two novel datasets (Large Grid and Movies paradigms). The former was specifically designed to segment EEG data in ocular events, allowing for the model training. In contrast, the Movies paradigm was used to validate the generalization of our approach in real-world situations. 

We performed extensive experiments on our novel and publicly available datasets (Visual Symbol Search \cite{kastrati2021eegeyenet} and Reading paradigms \cite{zuco2}). Based on the reported results in Table \ref{tab:architecture_performance}, we observed that our model outperforms the current state-of-the-art models by a large margin. Finally, we evaluated DETRtime in the different applications, namely the sleep staging task, initially performed with the U-time \cite{perslev2019utime} and SalientSleepNet \cite{salientsleepnet} models . Again, DETRtime outperformed current state of the art solutions for the sleep staging task, showing its generalization capabilities.

Additionally, our extensive experiments demonstrated that other deep learning models (CNNs, RNNs) outperformed standard machine learning techniques. We also observed that one of the main challenges was differentiating between vertical saccades and blinks because their signal trace is very similar \cite{kleifges2017blinker}. Furthermore, our dataset is heavily imbalanced and many samples exclusively contain fixations. This resulted in a higher F1 score for fixations than for saccades and blinks.

While DETRtime is an adaption of the original model (DETR), to the best of our knowledge, this is the first time when the deep learning-based methods developed initially for the  \textit{instance segmentation} (``object detection, bounding boxes'') have been applied in EEG data. Instead, all previous approaches adapted models designed for \textit{semantic segmentation} (most of which are convolution-based)\cite{perslev2019utime}. We believe this to constitute a fundamental difference in the methods used. Furthermore, we performed ablation studies showing that a separate segmentation pipeline is unnecessary in EEG data and experimented with several backbones specifically tailored for EEG data \cite{eegnet}.

 %Additionally, we demonstrate that other deep learning models (CNNs, InceptionTime, Xception, LSTMs, Pyramidal CNN) failed to learn this problem properly and argue about some of the main challenges of this task. 

\paragraph{Comparison to related work}

As described in Section \ref{sec:related-work}, the ocular events' detection is an active research topic in the applied machine learning community. Nonetheless, the previous results are not directly comparable to our setting. More precisely, we introduced the EEG segmentation task and an evaluation metric that unified several methods and problems addressed in related research. We classified signals to ocular events and segmented them with 2 ms time resolution. This way, we could also find each ocular event's onset (beginning) and offset (end).
On the other hand, \cite{behrens2010improved} only focused on detecting saccades with horizontal movement (left vs right) and finding other parameters of the saccade (such as acceleration). A closer work, \cite{bulling2011recognition} addressed the problem of detecting all three ocular events and reported an average of 0.761 score for precision and  0.705 for recall. In comparison, our DETRtime achieved an average of 0.936 precision score and 0.902 recall score.
Nevertheless, these results are also not directly comparable since \cite{bulling2011recognition} used an EOG dataset consisting of only 8 participants (as opposed to our datasets with in total 168 participants) and a different experimental paradigm while recording the data. Finally, \cite{pettersson2013algorithm} focused only on the saccade detection and did not follow the same setup as ours. 
It should be emphasized that our EEG segmentation model outperformed the current state-of-the-art solutions for the sleep staging task, providing a reliable tool for the time series segmentation.

\paragraph{Limitations and Future Work}
%We acknowledge that our models are trained on a dataset collected only on the large grid paradigm. This paradigm is well suited for the segmentation problem since it includes fixations in many different positions on the screen and saccades in many different directions. Nonetheless, showing the model ability to generalization on other datasets and domains remains to be done. 
In this work, we have observed different signal qualities across participants. Therefore, investigating other methods, such as fine-tuning the models on a single participant, is planned for future work. Furthermore, although we achieved a very high F1 score for the fixations detection, there is still room for improvement in detecting the other two ocular events. Additionally, our experiments revealed that the proposed task is much harder in naturalistic settings, making the contributed benchmarking task valuable for the ML community and fostering the development of new and more suitable methods.  

Finally, it is worth mentioning that the eye-tracking community is mainly focused on camera-based solutions. This work complements these research efforts showing that segmentation of ocular events can also be performed on EEG data with high accuracy. This opens the door to further research exploring whether combining these two modalities is beneficial.
Even though we provided evidence that DETRtime performs well in other time-series segmentation tasks such as sleep stage segmentation (see Appendix \ref{app:sleep-staging-task}), it also remains to experiment how DETRtime performs in the time-series datasets that do not necessarily stem out of EEG measurements. Nevertheless, the proposed framework is general, and the same architecture can be used for detecting different objects in time-series data.

\newpage

\bibliography{example_paper}
\bibliographystyle{icml2022}

%%%%%%%%%%%%%%%%%%%%%%%%%%%%%%%%%%%%%%%%%%%%%%%%%%%%%%%%%%%%%%%%%%%%%%%%%%%%%%%
%%%%%%%%%%%%%%%%%%%%%%%%%%%%%%%%%%%%%%%%%%%%%%%%%%%%%%%%%%%%%%%%%%%%%%%%%%%%%%%
% APPENDIX
%%%%%%%%%%%%%%%%%%%%%%%%%%%%%%%%%%%%%%%%%%%%%%%%%%%%%%%%%%%%%%%%%%%%%%%%%%%%%%%
%%%%%%%%%%%%%%%%%%%%%%%%%%%%%%%%%%%%%%%%%%%%%%%%%%%%%%%%%%%%%%%%%%%%%%%%%%%%%%%
\newpage
\appendix
\onecolumn
\section{Experiments with Segmentation Pipeline}
\label{app:segmentation pipeline}
\begin{figure*}[t]
    \centering
    \includegraphics[width=1\textwidth]{ 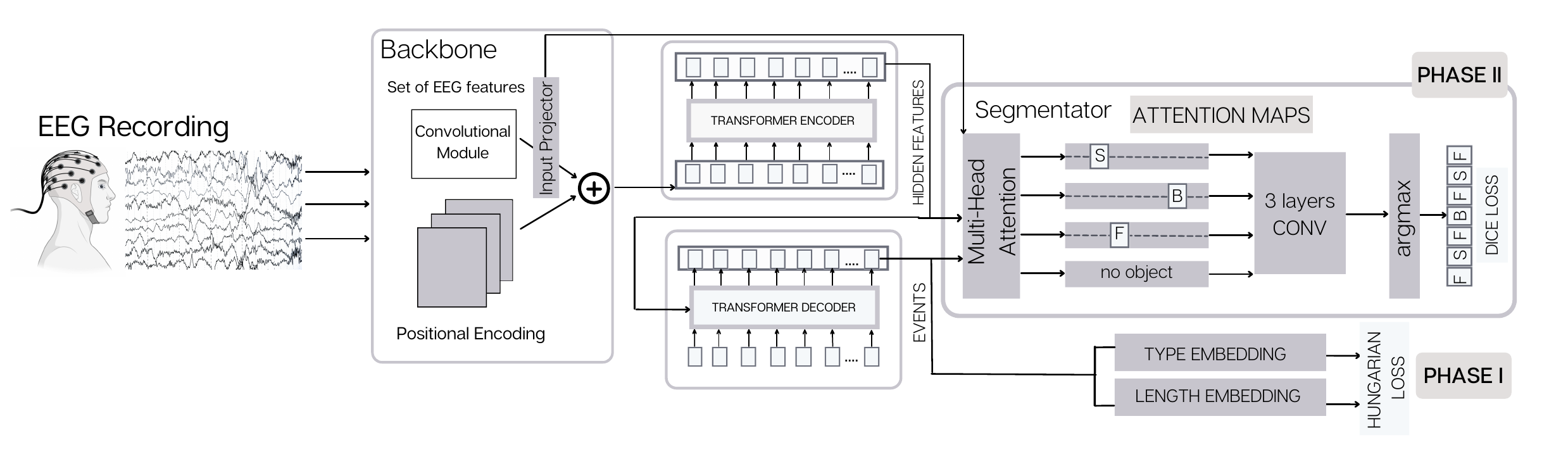}

    \caption{The experimental architecture described in Appendix \ref{app:segmentation pipeline} consists of a detection module and segmentation module. The detection module comprises a backbone model, positional embeddings, followed by a transformer and feed-forward networks that map the hidden features to the events and their length. The segmentation module that produces the segmentation masks for each detected event is composed of a multihead attention layer and a 3-layer convolutional module.}
    \label{fig:DETR_architecture2} 
    %\vspace{-0.4cm}
\end{figure*}

Instead of using the simple heuristic for predictions, we also tested whether a separate segmentation module would improve the results. Our model can be easily extended to perform segmentation in a 2-staged fashion, where we train a separate model on transformer feature maps, box embeddings, and inputs projections. We trained models adapted from the panoptic segmentation implemented in \cite{detr} (see Figure \ref{fig:DETR_architecture2}) as well as simpler CNN or multi-layer perceptron (MLP) models on our features. As we can see in Table \ref{tab:segmentation_pipeline_results}, none of these approaches brought any advantages, showing that the object detection and a simple heuristic, where we use argmax over overlapping events, is enough for the EEG segmentation task. In the following section, we explain our 2-stage approach. %Of these, an MLP on the hidden dimension of the box sequences performed best. Still, it was outperformed by a simple heuristic based on class confidences. 

\paragraph{Segmentation Module} As we can see in Figure \ref{fig:DETR_architecture2}, the segmentation module takes events features extracted from the decoder, the hidden features of the encoder and attends to a encoded signal from the backbone. The multi-headed attention module predicts heatmaps for each of the predicted events, which we further process with 3 convolution layers to produce the segmentation masks for each event.

%.  

\begin{table}[htb]
    \centering
    \begin{tabular}{lcr}
    \toprule
    Detr Backbone  &   F1 w/o Seg   &  F1 with Seg \\
    \midrule
    Pyramidal CNN &  \makebox{0.875} &  \makebox{0.874}  \\
    InceptionTime   &  \makebox{0.918} &  \makebox{0.907}  \\
    Xception  &  \makebox{0.873} &  \makebox{0.871}  \\
    \bottomrule
  \end{tabular}
  \caption{Segmentation Results.} 
  \label{tab:segmentation_pipeline_results}
\end{table}

We optimized the architecture in Figure \ref{fig:DETR_architecture2} in two different phases. 
\paragraph{Phase 1} In Phase 1, we train the original model using our segmentation objective. %We follow the same procedure as explained in Section \ref{sec:optimization}.
%Thus, this strategy allows precise refinement of our models weaknesses 

\paragraph{Phase 2} In Phase 2, we freeze the weights of the original model and train the additional segmentation module. The goal is to map the detected events (type, position, length) and learned feature maps of the transformer to the segmentation mask of the EEG data and thus classify each time point to a type of event. Here,  we have used the Adam optimizer again with a learning rate $\eta = 1e$-4. Similar to the baselines in Section \ref{app:deep learning baselines}, to produce the final segmentation masks, we used Dice Loss in Phase 2.

\section{Experiments on EEG Sleep Stage Segmentation Task}
\label{app:sleep-staging-task}
In order to further evaluate our model's performance, we provide evidence on a thoroughly researched segmentation task, namely EEG Sleep Stage Segmentation. Sleep staging segments a period of sleep into a sequence of phases providing the basis for most clinical decisions in sleep medicine \cite{Perslev2021}. We make use of the publicly available \textit{Sleep-EDF-153} dataset \cite{sleep-edf} and compare it to State-of-the-Art models in this research area. 

\begin{table*}[h]
\centering
    \scalebox{1.0}{
    \begin{tabular}{lcccccc}
    \toprule
      &    \multicolumn{6}{c}{Sleep-EDF-153}
     \\
    \midrule
   Model & W & N1 & N2 & N3 & REM & avg 
     \\
     \midrule
     % CNN LSTM & 0.85 & 0.31 & 0.58 & 0.58 & 0.86 & 0.54 & 0.71 & 0.71 & 0.75 & 0.44 & 0.36 & 0.52 & 0.91 & 0.00 & 0.66 & 0.00 & 0.00 & 0.31 \\
     SalientSleepNet & 0.93 & 0.54 & 0.86 & 0.78 & 0.86 & 0.795
     \\
     U-time & 0.92 & 0.51 & 0.84 & 0.75 & 0.80 & 0.76 
     \\
     DETRtime & 0.98 & 0.49 & 0.85 & 0.81 & 0.88 & \textbf{0.801}
     \\
    \bottomrule 
  \end{tabular}
  }
    \caption{Generalization on EEG Sleep Staging. Metric: Macro F1 score.}
    \label{tab:pretrained_results}
\end{table*}

\section{Hyperparameter Configuration of the Presented DETRtime Architecture}
\label{app:hyperparamsDETR}
In this section we give details about the hyperparameter configuration of the best found DETRtime model. The list can be found in Table \ref{tab:detr hyperparams}.

\begin{table}[h]
    \centering
    \scalebox{0.85}{
    \begin{tabular}{c|c}
        Hyperparameter & Value / Description \\
        \toprule
        Backbone & InceptionTime \\ 
        Learning Rate Backbone & 0.0001 \\ 
        Backbone Kernel Sizes & 16, 8, 4 \\
        Backbone Channels & 16 \\
        Backbone Depth & 6 \\ 
        Residual Connections & False \\
        \midrule
        Positional Embedding & Sine \\
        Transformer Encoding Layers & 6 \\
        Transformer Decoding Layers & 6 \\
        Dimension Feedforward & 2048 \\
        Hidden Dimension & 128 \\
        Dropout & 0.1 \\
        Number of Heads & 8 \\
        Number of Object Queries & 20 \\
        \midrule
        Cost Class & 1 \\
        Cost Bounding Box & 5 \\
        Cost Intersection over Union & 2 \\
        Bounding Box Loss Coefficient & 10 \\
        IoU Loss Coefficient & 2 \\
        No class Weight Coefficient & 0.3 \\
        Number of classes & 3 \\
        \midrule
        Sequence Length & 500 \\
        Random Seed & 42 \\
        \midrule
        Batch Size & 32 \\
        Weight Decay & 0.0001 \\
        Epochs & 200 \\
        Learning Rate Drop & Every 5 Epochs after 150 \\
        \bottomrule
    \end{tabular}
    }
    \caption{The hyperparameters of the best performing model: DETRtime.}
    \label{tab:detr hyperparams}
\end{table}

%lr_backbone=0.0001, batch_size=32, weight_decay=0.0001, epochs=300, lr_drop=200, clip_max_norm=0.1, backbone='inception_time', kernel_size=16, nb_filters=16, in_channels=129, out_channels=1, backbone_depth=6, use_residual=False, position_embedding='sine', back_channels=16, back_layers=6, enc_layers=6, dec_layers=6, dim_feedforward=2048, hidden_dim=128, dropout=0.1, nheads=8, num_queries=20, pre_norm=False, set_cost_class=1, set_cost_bbox=5, set_cost_giou=2, bbox_loss_coef=10.0, giou_loss_coef=2.0, eos_coef=0.3, num_classes=3, timestamps=500, data_path='/cluster/scratch/klebed/EEG/EEGEyeNet_experimental/data/ICML_participant_streams_thresh_150_seqlen_4000_margin_2/tensors', scaler=None, output_dir='./runs/run1', device='cuda', seed=42, resume='', start_epoch=0, eval=False, num_workers=2)
For DETRtime backbones, we tested the same CNN architectures that were used as baseline models. Accordingly, various hyperparameter configurations such as kernel sizes, channel numbers and module depth have been tested. Similarly, the transformer hyperparameters such as hidden dimensions and depth have been tested. Based on the number of events in a typical sample, we made the conscious decision to predict at most 20 events. This number is sufficiently high to cover almost all samples. The DETR loss weights have been adjusted to balance all loss components, thus ensuring smooth training progress. Little to no overfitting was experienced even after training over a large number of epochs.

\begin{center}
\begin{table*}[h]
    \centering
    \scalebox{0.9}{
        \begin{tabular}{@{}lllllllll@{}}
            \toprule
        {Parameter} & {CNN} & {Pyr. CNN} & {EEGNet} & {InceptionTime} & {Xception} & {LSTM} & {biLSTM} & {CNN-LSTM} 
        \\ 
        \midrule  
        {Depth} & \makebox{5} & \makebox{5} & \makebox{2} & \makebox{5} & \makebox{8} & 10 & 5 & \specialcell{5 CNN / \\ 3 LSTM} 
        \\
        {Number of filters} & \makebox{16} & \makebox{16} & \makebox{16/256} & \makebox{32} & \makebox{64} & - & - & 16
        \\
        {Kernel size} & \makebox{32} & \makebox{16} & \makebox{64} & \makebox{16} & \makebox{32} & - & - & 32
        \\
        {Max Pooling} & 2 & 2 & 2 & 3 & 2 & - & - & 2
        \\

        {Residual connections} & \makebox{True} & \makebox{False} & \makebox{False} & \makebox{True} & \makebox{True} & - & - & True 
        \\
        {Bottleneck size} & \makebox{-} & \makebox{-} &  \makebox{-} & \makebox{16} &  \makebox{-} & - & - & - 
        \\
        {Dropout rate} &  \makebox{-} &  \makebox{-} & \makebox{0.5} &  \makebox{-} &  \makebox{-} & 0.5 & 0.5 & 0.5 
        \\
        {LSTM Hidden Size} &  \makebox{-} &  \makebox{-} & - &  \makebox{-} &  \makebox{-} & 128 & 128 & 64 
        \\
        \midrule
        {Sequence Length} & 500 & 500 & 500 & 500 & 500 & 500 & 500 & 500 
        \\ 
        \midrule
        {Batch size} & \makebox{32} & \makebox{32} & \makebox{32} & \makebox{32} & \makebox{32} & 32 & 32 & 32
        \\
        {Epochs} & \makebox{50} & \makebox{50} & \makebox{50} & \makebox{50} & \makebox{50} & 50 & 100 & 50 
        \\
        {Early stopping patience} & \makebox{20} & \makebox{20} & \makebox{20} & \makebox{20} & \makebox{20} & 20 & 20
        \\
        {Optimizer} & Adam & Adam & Adam & Adam & Adam & Adam & Adam & Adam 
        \\ 
        {Learning Rate} & 1e-3 & 1e-3 & 1e-3 & 1e-3 & 1e-3 & 1e-3 & 1e-4 & 1e-3
        \\ 
        \bottomrule
        %& \makebox{0 \rpm 0.0} & \makebox{0 \rpm 0.0} & \makebox{0 \rpm 0.0} & \makebox{0 \rpm 0} \\   
        \bottomrule
        \end{tabular}
    }
\caption{Hyperparameters of the Deep Learning Models. EEGNet and U-Net both report the respective number of filters for all their blocks.}
\label{table:hyper_classification}
\end{table*}
\end{center}

\begin{center}
\begin{table*}[h]
    \centering
    \scalebox{0.9}{
        \begin{tabular}{@{}lllllllllll@{}}
            \toprule
        {Parameter} &  {SalientSleepNet} & {U-Time}
        \\ 
        \midrule  
        {Depth} & 7 & \makebox{7}
        \\
        {Number of filters} & \specialcell{128, 256, 512, \\ 1024, 512, 256, \\  128} & \specialcell{256, 512, 1024 \\ 2048 1024, 512, \\ 256}
        \\
        {Kernel size} & 5 & 5 
        \\
        {Max Pooling} & 2 & 4/3/2 
        \\
        {Upsampling} & 2 & 2/3/4 
        \\
        {Residual connections} & True & \makebox{True}
        \\
        \midrule
        {Sequence Length} & 500 & 500 
        \\ 
        \midrule
        {Batch size} & 32 & 32 
        \\
        {Epochs} & 50 & 100 
        \\
        {Early stopping patience} & 20 & 20
        \\
        {Optimizer} & Adam & Adam 
        \\ 
        {Learning Rate} & 1e-3 & 1e-5 &
        \\ 
        \bottomrule
        \bottomrule
        \end{tabular}
    }
\caption{Hyperparameters of the State-of-the-Art Segmentation Models. We report the respective number of filters for all their blocks.}
\label{table:hyper-state-of-the-art}
\end{table*}
\end{center}

\begin{center}
    \begin{table*}[h]
    \centering
    \scalebox{1}{
        \begin{tabular}{@{}lllll@{}}
            \toprule
        {Parameter} & {kNN} & {Decision Tree} & {Random Forest} & {Ridge Classifier}
        \\ 
        \midrule  
        {Number of Neighbors} & \makebox{5} & \makebox{-} & \makebox{-} & \makebox{-} 
        \\
        {Weights} & \makebox{Uniform} & \makebox{-} & \makebox{-} & \makebox{-} 
        \\
        {Criterion} & \makebox{-} & \makebox{Gini} & \makebox{Gini} & \makebox{-} 
        \\
        {Max Depth} & \makebox{-} & \makebox{None} & \makebox{None} & \makebox{-} 
        \\
        {Number of Estimators} & \makebox{-} & \makebox{-} & \makebox{150} & \makebox{-} 
        \\
        {Alpha} & \makebox{-} & \makebox{-} & \makebox{-} & \makebox{1.0} 
        \\
        {Tol} & \makebox{-} & \makebox{-} & \makebox{-} & \makebox{1e-3} 
        \\
        \bottomrule
        %& \makebox{0 \rpm 0.0} & \makebox{0 \rpm 0.0} & \makebox{0 \rpm 0.0} & \makebox{0 \rpm 0} \\   
        \bottomrule
        \end{tabular}
    }
    %\vspace{0.3cm}
\caption{Hyperparameters of the Standard Machine Learning Models from scikit-learn.}
\label{table:hyper_classification2}
\end{table*}
\end{center}

\section{Hyperparameter Configurations of the Baseline Models}
\label{app:hyperparamsBaseline}
In Table \ref{table:hyper_classification} and Table \ref{table:hyper_classification2}, we report the hyperparameters of all baseline models. Hyperparameters were optimized on the Large Grid paradigm dataset.

\section{Detailed Baseline Model Overview}
\label{app:detailed model overview}
%In this section briefly cover all models used for establishing our baselines.

\subsection{Naive Baselines}
\label{app: naive baselines}
In order to assess the performances of our model suite, we provide three types of naive predictions: sampling uniformly at random from the three classes (fixation, saccade, blink), sampling from the prior label distribution, and always predicting the most frequent class (fixation). For details about the label distributions we refer to Appendix \ref{app:event details}

\subsection{Standard Machine Learning Baselines}
\label{app:standard machine learning models detailed}
We briefly introduce our standard machine learning baseline models. For all models explored here we make use of the \textit{scikit-learn} \cite{scikit-learn} implementation. 

\subsubsection{k Nearest Neighbors}
\label{sec:kNN}
The k-nearest neighbors (kNN) algorithm is a non-parametric method that can be used for classification and regression. The inputs used are the k-closest training samples in the data set. Using the algorithm for classification, we choose the predicted class based on a majority vote of the k nearest neighbors. In regression, the output is the average of the values of the k nearest neighbors. 

\subsubsection{Decision Tree}
\label{Decision Tree}
Decision Tree is a non-parametric supervised learning method that can be used for both classification and regression. The core idea of the model is to predict target values by learning simple decision rules based on input data features. A tree can be seen as piecewise constant approximation. 

\subsubsection{Random Forest}
\label{Random Forest}
Random forests are an ensemble learning method for classification and regression that operates by constructing multiple decision trees at training time. For classification tasks, the output of the random forest is the class selected by most trees. When performing regression, the mean of average of the predictions of the individual trees is returned. 

\subsubsection{Ridge Classifier}
\label{Ridge Classifier}
Ridge regression addresses some of the problems of Ordinary Least Squares by imposing a penalty on the size of the coefficients. The ridge coefficients minimize a penalized residual sum of squares objective. The Ridge Classifier is the classifier variant of Ridge Regression. The classifier first converts binary targets to $\{-1,1\}$ and then treats the problem as a regression task, optimizing the Ridge Regression objective. In our case of multi-class classification, the problem can be reformulated from multi-output regression, where the predicted class is found by computing the argmax of output values.

\subsection{Established Deep Learning Baselines}
\label{app:deep learning baselines}
Each of the following models has an input sequence of size (N,C=128) and predicts a label sequence of size (N,1), where N is the length of the EEG data stream and C is the number of channels (electrodes). Since experiments with CrossEntropy and Focal Loss \cite{focalloss} turned out to perform poorly, we decided to experiment with Dice Loss, a region-based loss function classically used for semantic segmentation \cite{li2020dice}. The Dice Loss is based on the Sørensen–Dice Coefficient that measures the normalized intersecting area of prediction and ground truth. 

All models in our baseline model suite were trained with Dice loss. Normalized class weights were assigned to the loss function, that are equal to the inverse of the relative occurrence of the respective class. In addition to that, we made use of the \textit{biased sampling} technique covered in Section \ref{sec:dataset-imbalance}. In the remainder of this section, we will introduce our baseline models. 

\subsubsection{Convolutional Neural Network (CNN)}
\label{app:cnn}
This is the most basic model in our collection of deep learning models. We implement a standard convolutional neural network with one-dimensional convolutional filters and additional residual connections every three convolutional blocks. Each of the blocks consists of such a 1D-convolution, a batch-normalization layer, and applying the ReLU activation function to the output followed by a max pooling layer. In the configuration used for the reported results, we use 16 filters of kernel size 32 (time samples), and the pooling operation uses a kernel size of 2 with a stride of 1.

\subsubsection{Pyramidal CNN}
\label{app:pyramidal-cnn}
This model implements a classical CNN where we stack 8 CNN blocks with sizes that follow the shape of an inverted pyramid. Each block consists of the same modules as in the CNN describe above, with the difference that the number of filters is a multiple of 16 and grows with depth. This means we have 16 filters in the first layer, 32 in the second and so on. We used a kernel size of 16 for the convolutional layers. As in the CNN, the convolution is followed by a batch normalization layer and a ReLu activation. Finally, a max-pooling layer is applied with a kernel size of size 2 and a stride of size 1. Note that in this architecture we did not use residual connections due to the different shape of the output compared to the input of a block. 

\subsubsection{EEGNet}
\label{app:eegnet}
The EEGNet architecture is a convolutional neural network developed for Brain-Computer Interfaces \cite{eegnet}. This model performs a temporal convolution to learn frequency filters and then performs a depthwise convolution to learn frequency-specific spatial filters followed by a separable convolution which should learn a temporal summary for each individual feature map. Finally, to mix the results of these feature maps, a pointwise convolution is applied. 

\subsubsection{InceptionTime}
\label{app:inception}
This models implementation is based on \cite{inception}, an adaption of the Inception-v4 architecture \cite{szegedy2017inception} for time series classification. We build this model from 9 blocks and skip connections every three layers. Each block consists of an \textit{InceptionTime} module, which takes as input 64 channels and performs a $1 x 1$ bottleneck convolution that produces a feature map with 16 channels. The reason for reducing the width is the following: we then pass this feature map through three different convolution operators with 16 filters each, using kernel sizes of 16, 8, and 4. What follows is a max-pooling layer with a kernel size of 3. The four resulting feature maps are then concatenated to form the layer's output, which has, as the input, again 64 channels. The key idea is that the network itself offers a variety of possible convolution operators at each layer. The bottleneck convolution at the beginning reduces the dimensions and targets to reduce the increased amount of computation necessary due to the 4 parallel convolutions. 

\subsubsection{Xception}
\label{app:xception}
As part of our baselines we also provide a model based on the Xception architecture proposed by \cite{xception}. This model has the same structure as the CNN, with 12 layers and residual connections. Each layer contains a 1D depthwise separable convolution \cite{xception}, with 64 filters and a kernel size of 32, followed by a batch normalization layer and ReLu activation.

\subsubsection{LSTM}
\label{app:lstm}
A recurrent neural network (RNN) \cite{rnn-alex} is a type of artificial neural network which uses sequential data or time series data. Unfortunately, RNNs are incapable of tracking long-range dependencies. The Long-Short-Term-Memory (LSTM) cells \cite{lstm} deal with these problems by introducing new gates, such as input and forget gates, which allow for a better control over the gradient flow and enable better preservation of long-range dependencies. LSTM-based models have shown great performance in sequence processing tasks. Unlike the standard LSTM, in the bidirectional LSTM (biLSTM) the input flows in both directions. Therefore, it’s capable of utilizing information from both sides. As LSTM-based models have shown great success in sequence processing, we include both LSTM and biLSTM implementations into our baseline model suite, where we make use of the \textit{PyTorch} \cite{pytorch_neurips_2019_9015} standard implementations. For both model variants we chose a hidden size of 128 and a dropout rate of 0.5.

\subsubsection{CNN-LSTM}
\label{app:cnn-lstm}
The CNN Long Short-Term Memory Network (CNN-LSTM) is an LSTM architecture specifically designed for sequence prediction problems. Successful use cases can be found in the domain of visual recognition and description \cite{cnn-lstm}. The CNN-LSTM architecture involves using CNN layers for feature extraction on input data combined with LSTMs to support sequence prediction. We stack 5 CNN blocks as described in Section \ref{app:cnn}, followed by a unidirectional LSTM block of depth 3 with a hidden size of 64 and a dropout rate of 0.5.

%\paragraph{GazeNet}
%\label{app:gazenet}
%The GazeNet architecture was inspired by by Deep Speech 2, an end-to-end speech recognition neural network. Our network is a re-implementation of \cite{zemblys2018gazeNet} where the complexity was increased for our purposes. It is an CNN-RNN architecture where a set of convolutional layers is followed by recurrent layers. In our implementation, we stack 6 convolutional layers before 4 bi-directional recurrent layers. The generated feature sequence is then passed through an additional convolutional layer and fully connected layer to produce the final output.
%The idea of this architecture is to have the convolutional layers extract features while the recurrent layers detect the corresponding event sequences.
%We stack 3 convolutional layers, followed by five bi-directional recurrent layers as well as a downsampling (convolution) and fully connected layer to compute the final output. 
%The idea is to first extract features with convolutional layers, while recurrent layers model event sequences and are responsible for detecting onsets and offsets of fixations, saccades and blinks.

\subsection{State-of-the-Art Deep Learning Baselines}
\label{app: state-of-the-art-baselines}

\subsubsection{U-Time}
\label{appsec:utime}
U-time is an approach to time series segmentation proposed in \cite{perslev2019utime}. It is based on the U-Net CNN architecture originally designed for semantic segmentation of biomedical images \cite{DBLP:journals/corr/RonnebergerFB15}. The idea is to first contract the input through multiple convolutional and pooling layers before upsampling the features to original resolution size again through layers of convolutional and upsampling layers. Thus, the architecture consists of a layer of encoding blocks followed by decoding blocks, effectively yielding a U-shape architecture. In addition, skip connections are added between matching encoder and decoder blocks. The contraction allows for coverage over larger patches in the sequence, while the skip connections improve localization accuracy of detected patches. We use blocks of two consecutive convolutional layers, each followed by an activation and batch normalization layer. In addition, each encoding block has a final maxpooling layer, while decoding blocks have an initial upsampling block respectively. The overall architecture consists of 3 encoding blocks, a bottleneck convolutional layer and 3 decoding blocks. Before each decoding block the feature outputs of its counterpart encoding block are added to the output of the preceeding block. The hyperparameters can be found in Table \ref{table:hyper-state-of-the-art}.
% For further processing, we also added a CRF layer to refine our predictions as it has already proven itself successful in semantic segmentation \cite{deeplab_crf}. %U-Net was then trained either directly using the various loss functions or using sequence likelihood in case of the CRF implementation. 

\subsubsection{SalientSleepNet}
\label{appsec-salientsleep}
SalientSleepNet is a multimodal salient wave detection network for sleep staging. The fully convolutional network is based on the $U^2$-Net architecture that was originally proposed for salient object detection in computer vision. It is mainly composed of two independent $U^2$-like streams to extract the salient features from multimodal data, respectively. We summarize the five key ideas of the architecture \cite{salientsleepnet}: 
1) Develop a two-stream $U^2$-structure to capture the salient waves in EEG and EOG modalities. 2) Design a multi-scale extraction module by dilated convolution with different scales of receptive fields to learn the multi-scale sleep transition rules explicitly. 3) Propose a multimodal attention module to fuse the outputs from EEG and EOG streams and strengthen the features of different modalities which make greater contribution to identify certain sleep stage. 4) Improve the traditional pixel-wise (point-wise) classifier in computer vision into a segment-wise classifier for sleep signals. 5) Employ a bottleneck layer to reduce the computational cost to make the overall model lightweight. The hyperparameters can be found in Table \ref{table:hyper-state-of-the-art}.

\section{EEG preprocessing pipeline}
\label{appsec-eeg}

The EEG preprocessing was conducted with the open-source MATLAB toolbox pipeline Automagic \cite{pedroni2019automagic}, which combines state-of-the-art EEG preprocessing tools into a standardized and automated pipeline. The EEG preprocessing consisted of the following steps: First, bad channels were detected by the algorithms implemented in the EEGlab plugin \texttt{clean\_rawdata}.\footnote{\url{http://sccn.ucsd.edu/wiki/Plugin\_list\_process}} Detected bad channels were automatically removed and later interpolated using a spherical spline interpolation. Subsequently, residual bad channels were excluded if their standard deviation exceeded a threshold of $25 \mu V $. Very high transient artifacts ($> \pm100 \mu V$) were excluded from calculating the standard deviation of each channel. Next, line noise artifacts were removed by applying Zapline \citep{de2020zapline}. However, if this resulted in a significant loss of channel data ($>$ 50\%), the channel was removed from the data.  
Due to the poor data quality and missing parts of the data, we removed 14 participants' recordings from the dataset. Therefore, the final sample used for experiments consists of recordings from 168 subjects.

\section{Details on Ocular Events}
\label{app:event details}
In this section we provide detailed information about the events occurring in the four datasets we provide, namely \textit{Movie Watching-}, \textit{Reading-}, \textit{Visual Symbol Search-} (VSS), as well as \textit{Large Grid paradigm}. Visualisations of the experimental setups can be found in Section \ref{sec:stimuli-and-experimental-design}.

\subsection{Movie Watching paradigm}
\label{app:movie}
\begin{figure}[h]
    \centering
    \begin{subfigure}{0.5\textwidth}
    \includegraphics[width=\linewidth]{ 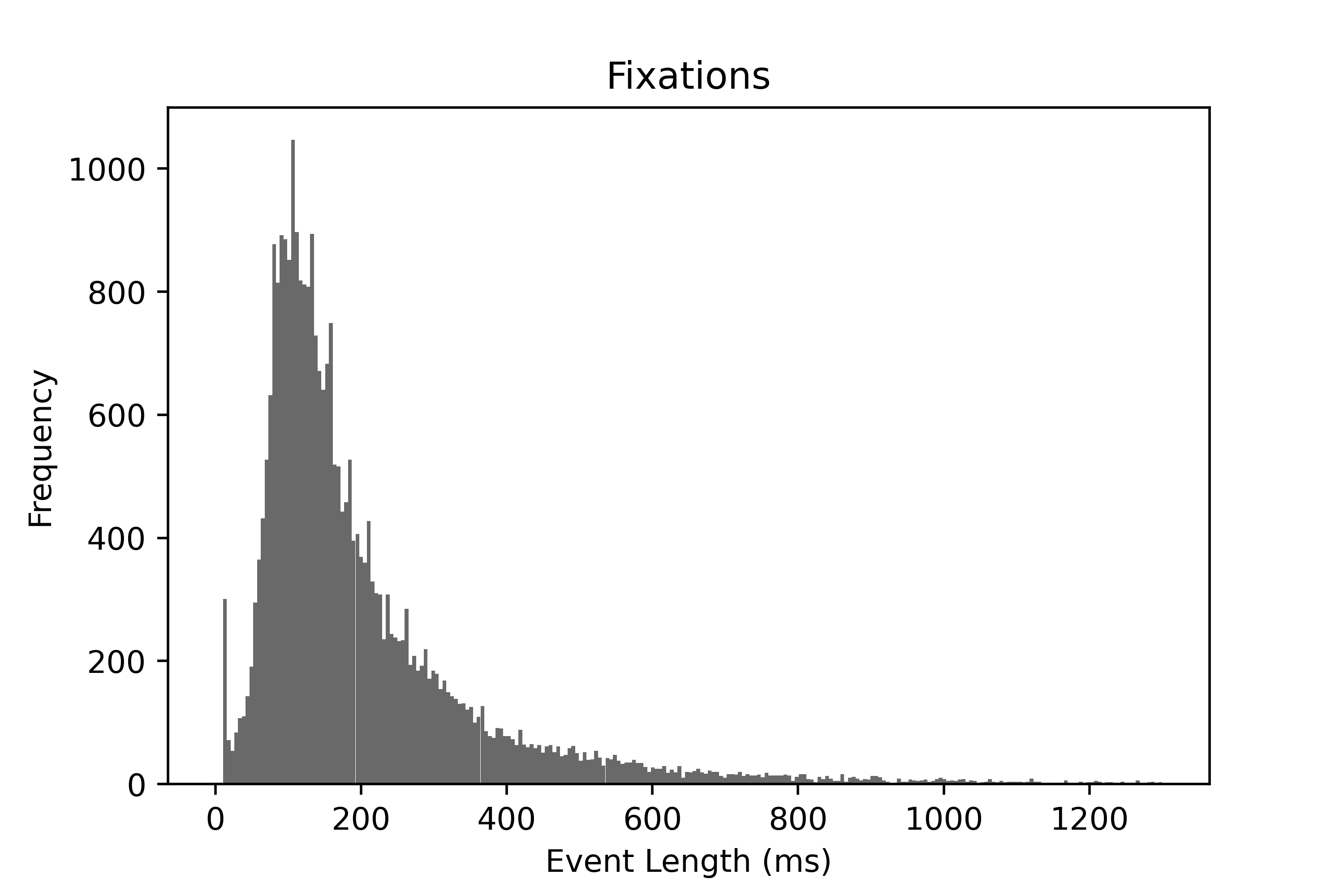}
    \caption{Distribution of the fixation length}
    \end{subfigure}%
    \begin{subfigure}{.5\textwidth}
    \includegraphics[width=\linewidth]{ 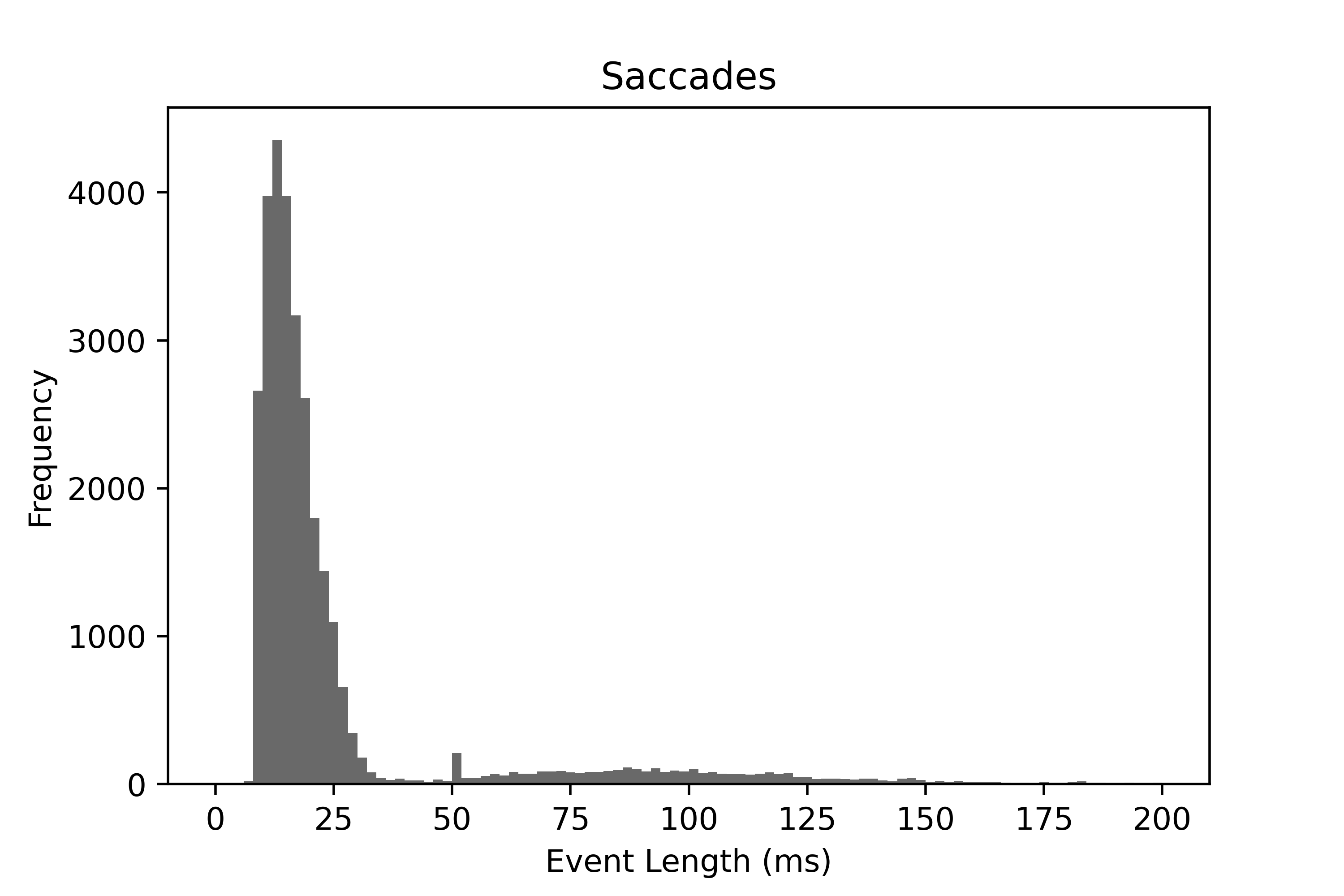}
    \caption{Distribution of the saccade length}
    \end{subfigure}
    \begin{subfigure}{.5\textwidth}
    \includegraphics[width=\linewidth]{ 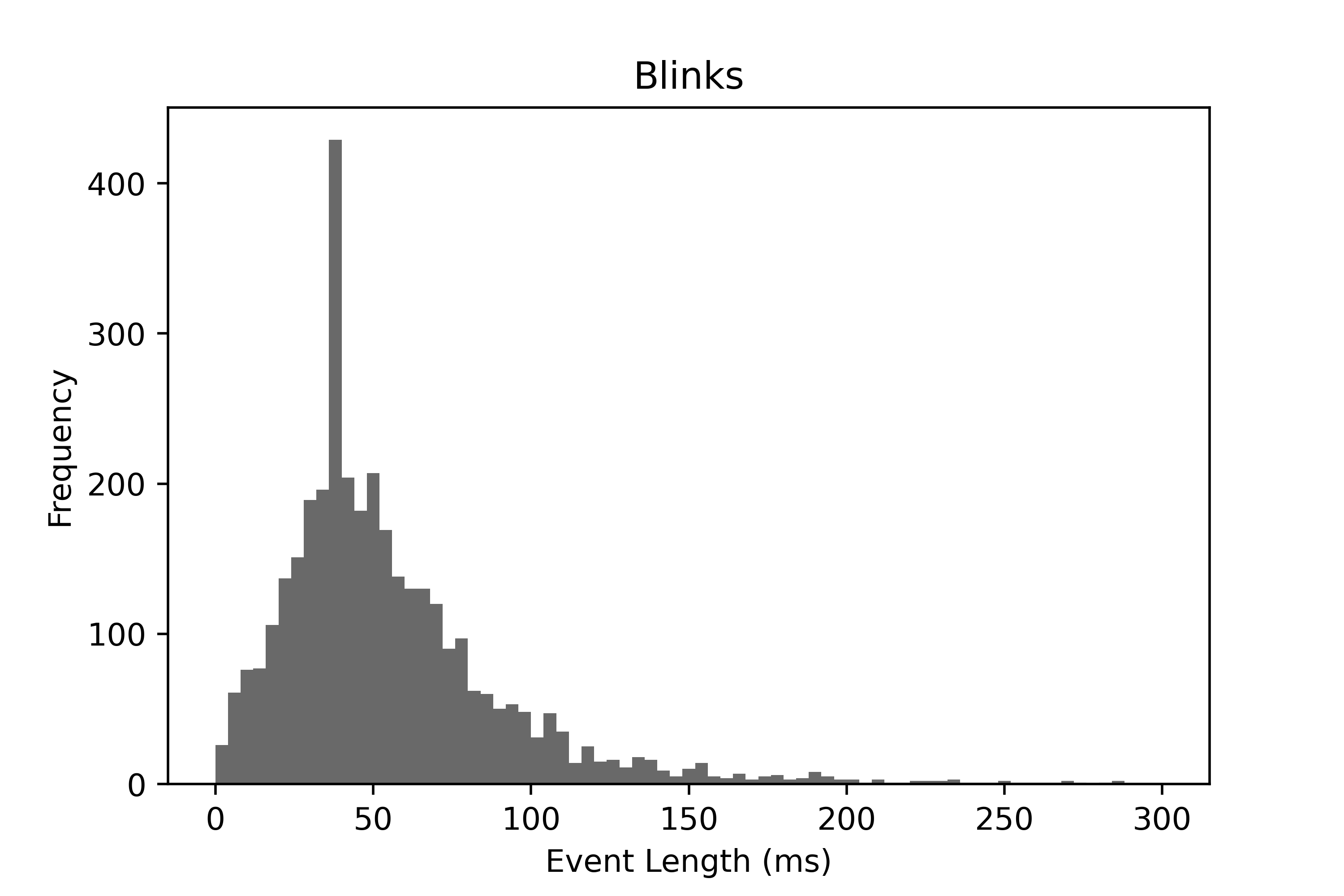}
    \caption{Distribution of the blink length}
    \end{subfigure}
    \caption{Distribution of the length of each ocular event in the \textbf{Movie Watching paradigm}: fixations, saccades and blinks.}
    \label{fig:histogram-movie}
\end{figure}

In the Movie Watching paradigm, we observe the following label distribution: fixation 86.49\%, saccade 10.77\%, blink: 2.74\%. The fixations have an average length of 216 time samples (432ms) with a standard deviation of 227 (454ms). In contrast to the Large Grid paradigm, the mean value is a common value of the distribution. In the saccade class, we have an average events' length of 27 (54ms) and a standard deviation of 39 (78ms). Blink events have an average length of 59 (118ms) with a standard deviation of 72 (144ms). Details of the distributions can be found in Figure \ref{fig:histogram-movie}. Event lengths are given in terms of measurement points (i.e. 2ms per point). 

\subsection{Natural Reading paradigm}
\label{app:natural-reading}
\begin{figure}[h]
    \centering
    \begin{subfigure}{0.5\textwidth}
    \includegraphics[width=\linewidth]{ 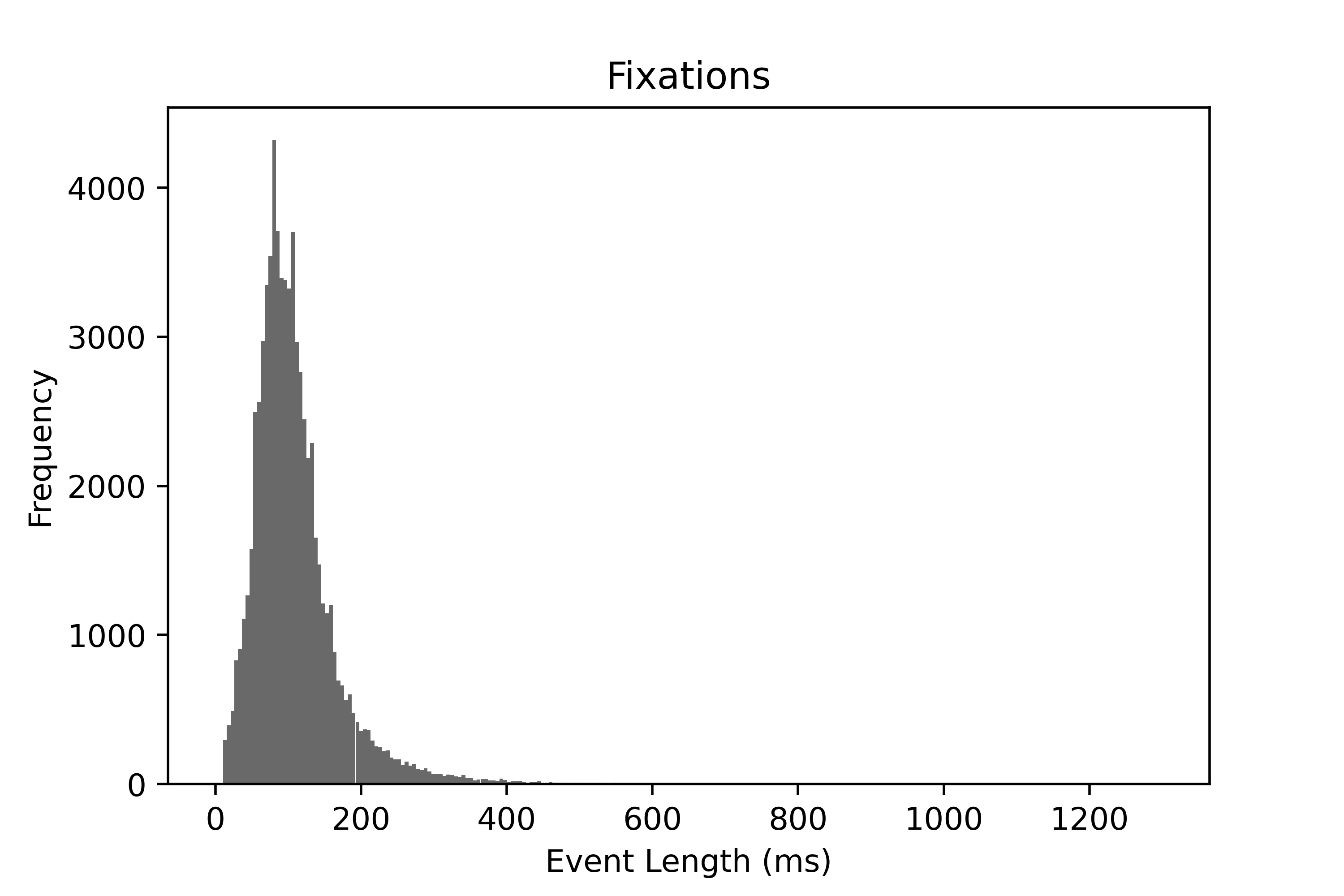}
    \caption{Distribution of the fixation length}
    \end{subfigure}%
    \begin{subfigure}{.5\textwidth}
    \includegraphics[width=\linewidth]{ 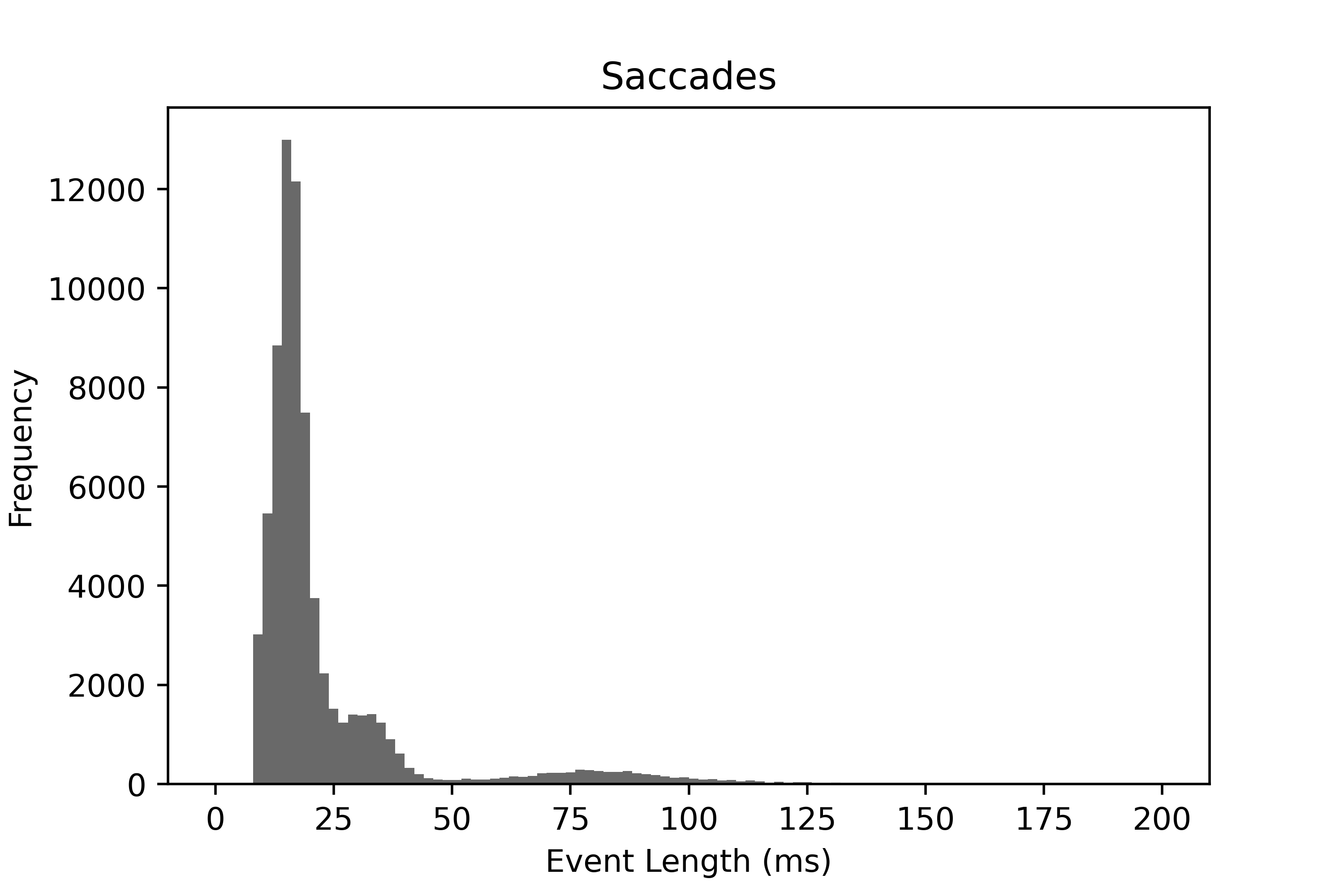}
    \caption{Distribution of the saccade length}
    \end{subfigure}
    \begin{subfigure}{.5\textwidth}
    \includegraphics[width=\linewidth]{ 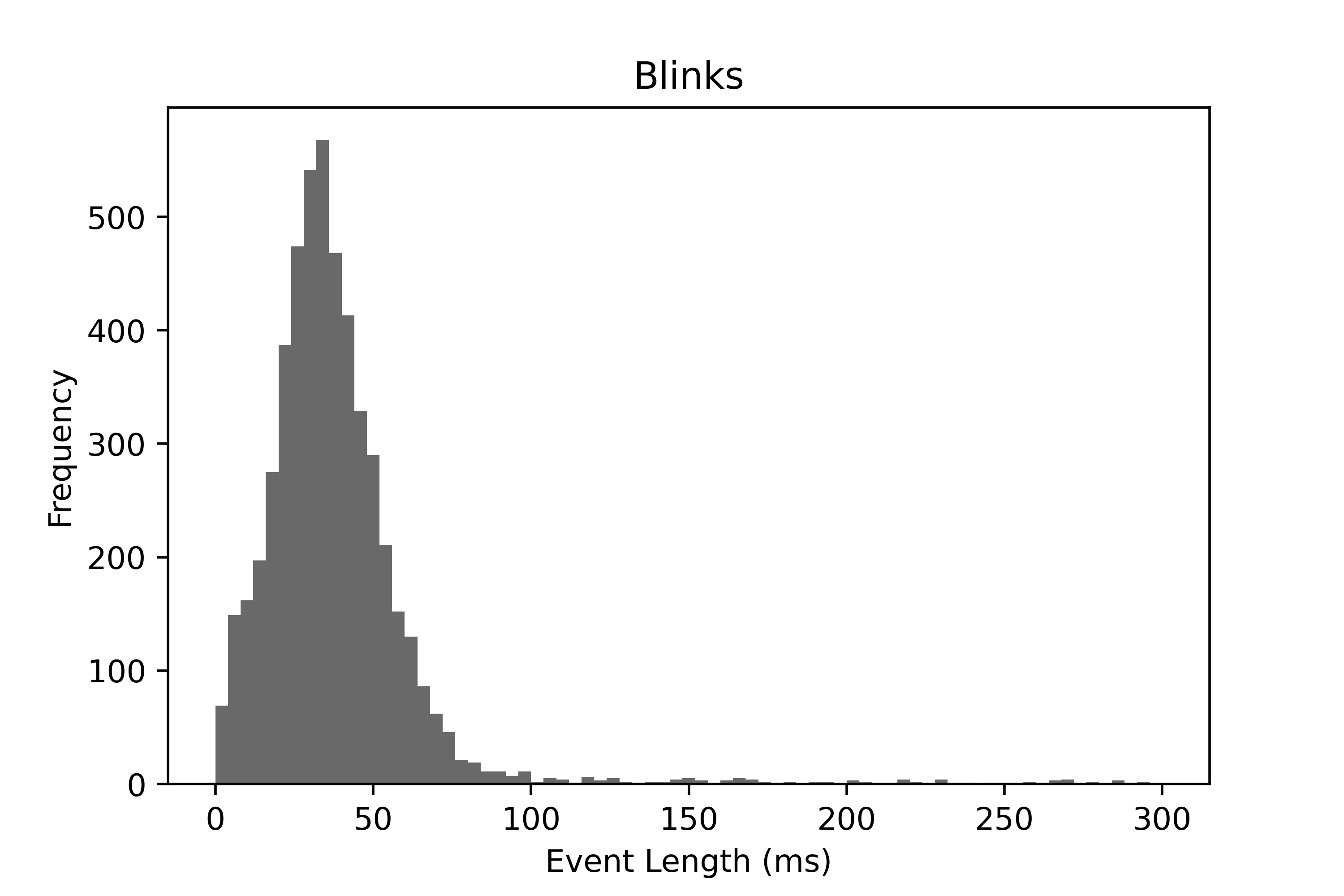}
    \caption{Distribution of the blink length}
    \end{subfigure}
    \caption{Distribution of the length of each ocular event in the \textbf{Natural Reading paradigm}: fixations, saccades and blinks.}
    \label{fig:histogram-zuco}
\end{figure}

In the Natural Reading paradigm, we observe the following label distribution: fixation 79.50\%, saccade 17.63\%, blink: 2.87\%. The fixations have an average length of 110 time samples (220ms) with a standard deviation of 63 (126ms). In the saccade class, we have an average events' length of 24 (48ms) and a standard deviation of 49 (98ms). Blink events have an average length of 54 (108ms) with a standard deviation of 167 (334ms). Details of the distributions can be found in Figure \ref{fig:histogram-zuco}. Event lengths are given in terms of measurement points (i.e. 2ms per point). 

\subsection{Visual Symbol Search paradigm}
\label{app:vss}
\begin{figure}[h]
    \centering
    \begin{subfigure}{0.5\textwidth}
    \includegraphics[width=\linewidth]{ 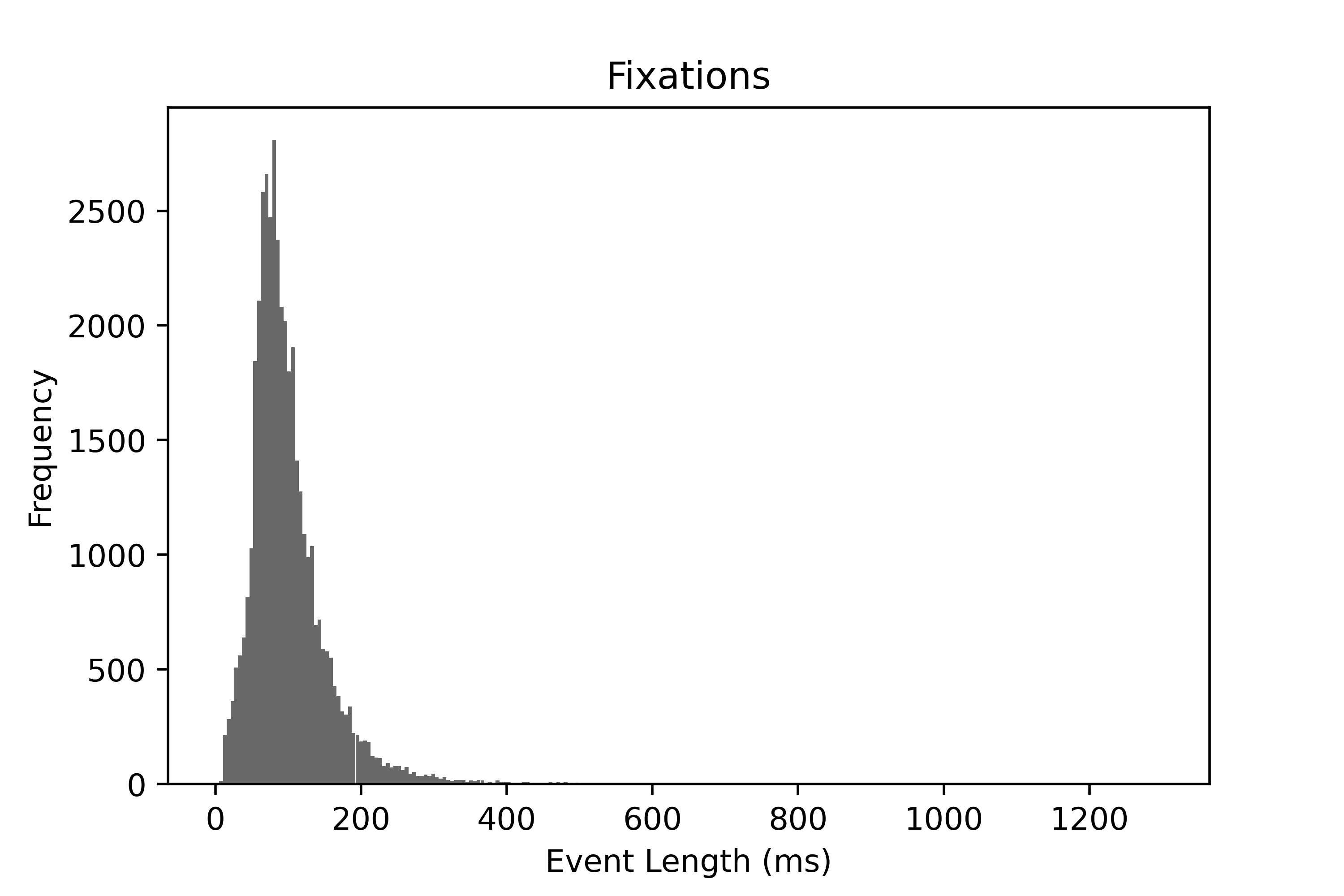}
    \caption{Distribution of the fixation length}
    \end{subfigure}%
    \begin{subfigure}{.5\textwidth}
    \includegraphics[width=\linewidth]{ 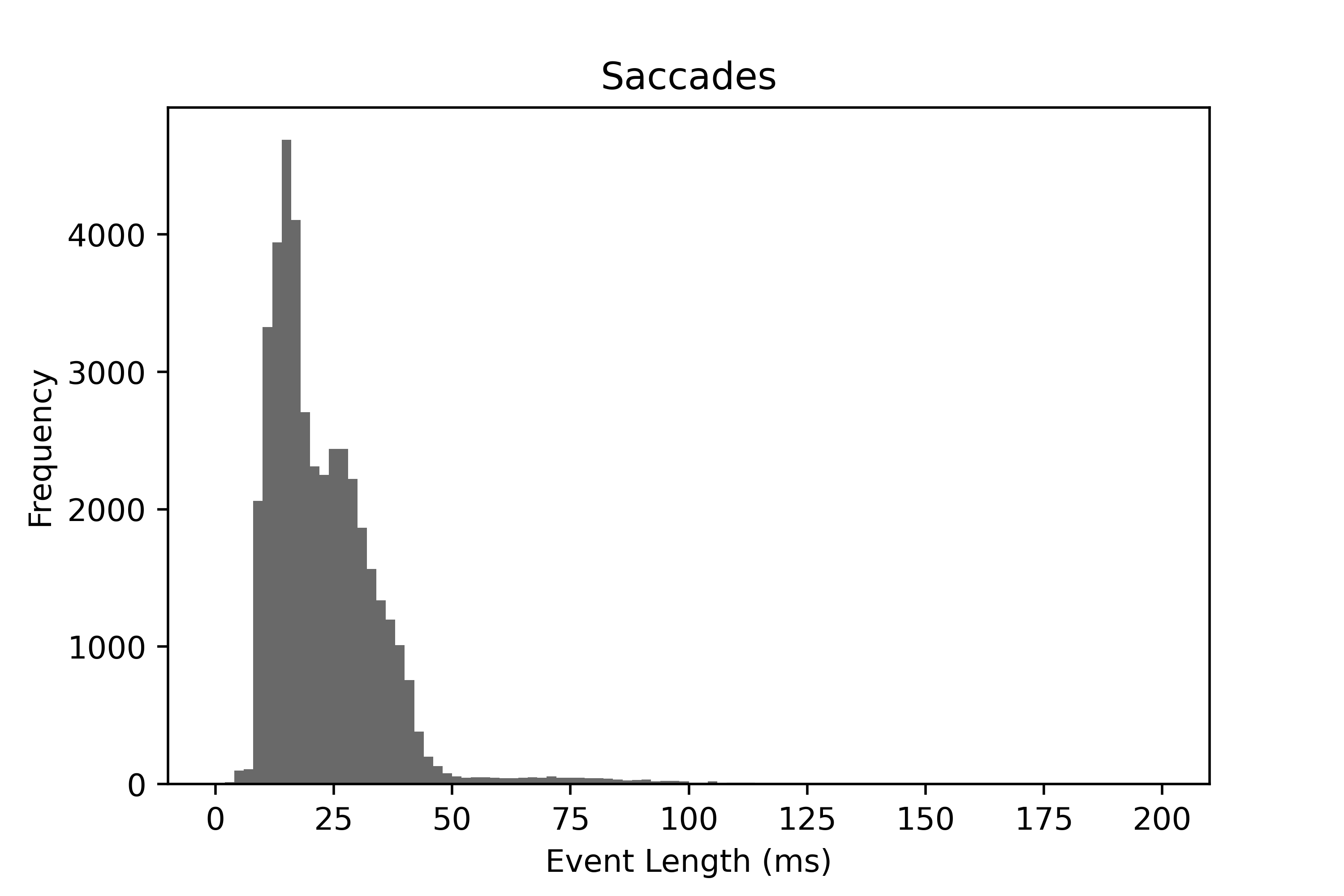}
    \caption{Distribution of the saccade length}
    \end{subfigure}
    \begin{subfigure}{.5\textwidth}
    \includegraphics[width=\linewidth]{ 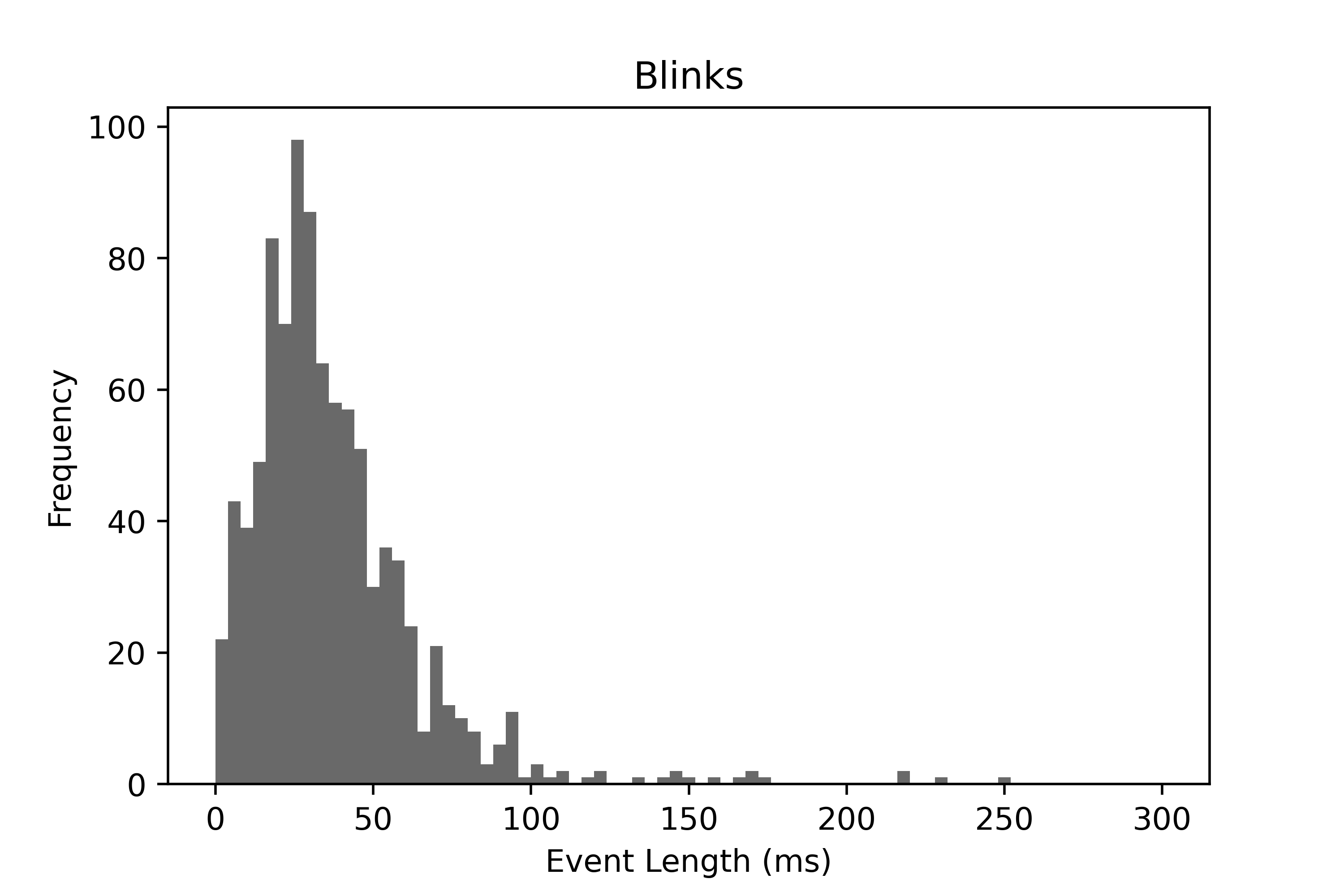}
    \caption{Distribution of the blink length}
    \end{subfigure}
    \caption{Distribution of the length of each ocular event in the \textbf{Visual Symbol Search paradigm}: fixations, saccades and blinks.}
    \label{fig:histogram-vss}
\end{figure}

In the Natural Reading paradigm, we observe the following label distribution: fixation 80.96\%, saccade 18.38\%, blink: 0.66\%. The fixations have an average length of 100 time samples (200ms) with a standard deviation of 56 (112ms). In the saccade class, we have an average events' length of 23 (46ms) and a standard deviation of 14 (28ms). Blink events have an average length of 37 (74ms) with a standard deviation of 28 (56ms). Details of the distributions can be found in Figure \ref{fig:histogram-vss}. Event lengths are given in terms of measurement points (i.e. 2ms per point). 

\subsection{Large Grid paradigm}
\label{app:large-grid}
\begin{figure}[h]
    \centering
    \begin{subfigure}{0.5\textwidth}
    \includegraphics[width=\linewidth]{ 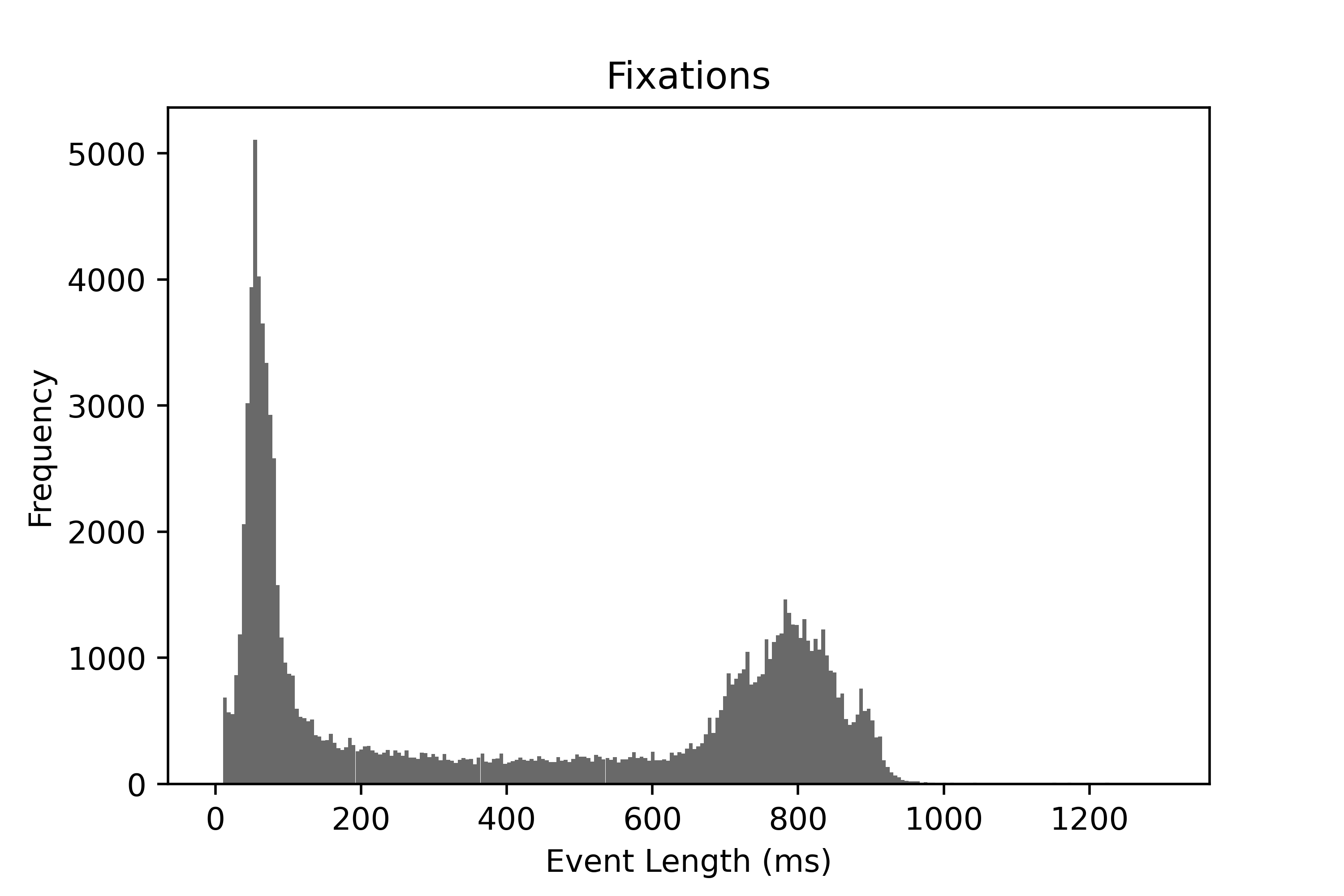}
    \caption{Distribution of the fixation length}
    \end{subfigure}%
    \begin{subfigure}{.5\textwidth}
    \includegraphics[width=\linewidth]{ 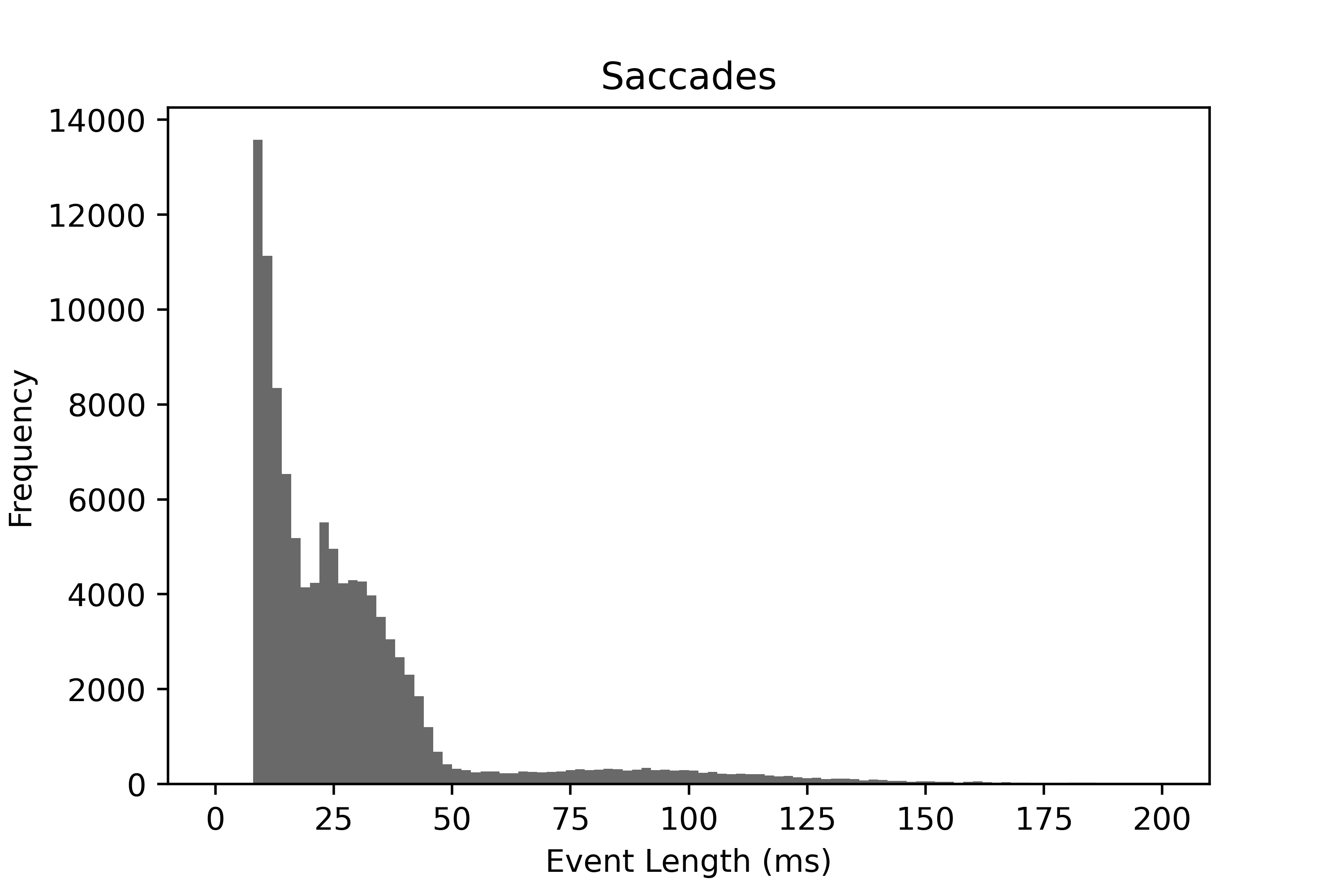}
    \caption{Distribution of the saccade length}
    \end{subfigure}
    \begin{subfigure}{.5\textwidth}
    \includegraphics[width=\linewidth]{ 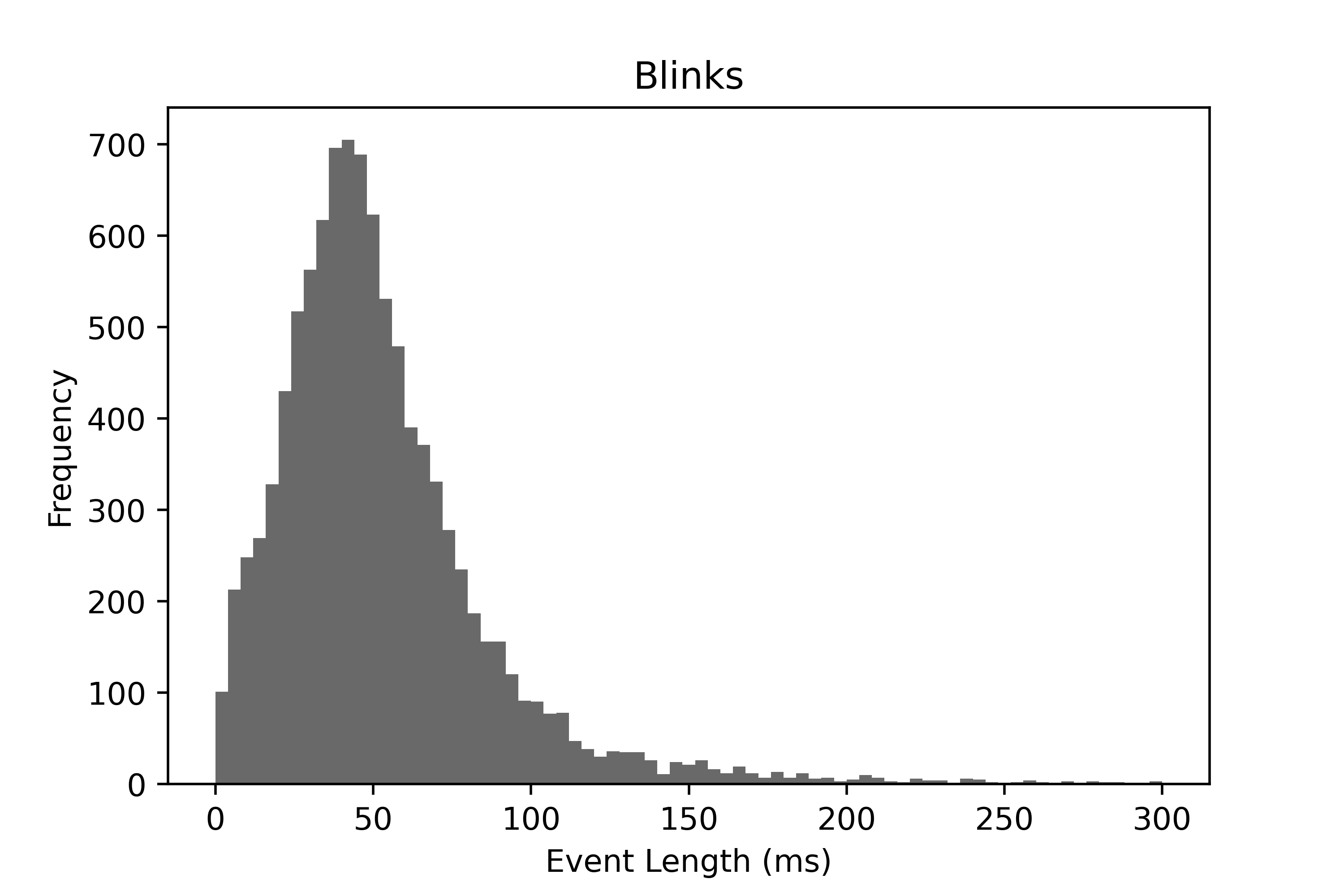}
    \caption{Distribution of the blink length}
    \end{subfigure}
    \caption{Distribution of the length of each ocular event in the \textbf{Large Grid paradigm}: fixations, saccades and blinks.}
    \label{fig:histogram-large-grid}
\end{figure}

In the Large Grid paradigm, we observe the following label distribution: fixation 92.26\%, saccade 6.59\%, blink: 1.15\%. The fixations have an average length of 421 time samples (842ms) with a standard deviation of 359 (718ms). It's noteworthy that the mean value is here almost never observed. In the saccade class, we have an average events' length of 30 (60ms) and a standard deviation of 35 (70ms). Blink events have an average length of 56 (112ms) with a standard deviation of 63 (126ms). Details of the distributions can be found in Figure \ref{fig:histogram-large-grid}. Event lengths are given in terms of measurement points (i.e. 2ms per point).

\end{document}